\documentclass[11pt,numbers,authoryear,round]{elsarticle}

\usepackage[english]{babel}

\usepackage[letterpaper,top=2cm,bottom=2cm,left=3cm,right=3cm,marginparwidth=1.75cm]{geometry}

\usepackage{amsmath}
\usepackage{graphicx}
\usepackage[colorlinks=true, allcolors=blue]{hyperref}
\usepackage{listings}
\usepackage{xcolor}
\usepackage{float}
\usepackage{titlesec}

\titleformat{\paragraph}[runin]
  {\normalfont\normalsize\bfseries}
  {}
  {0pt}
  {}

\titlespacing*{\paragraph}
  {2em}    
  {0pt}    
  {0.5em}  

\begin{document}

\begin{frontmatter}

\title{SimSkill: A Self-Evolving LLM Agent for Skill and Knowledge Accumulation in Traffic Simulation}

\author[a]{Qi Liu}
\author{Qinzheng Wang$^{a,\ast }$}
\author{Can Li$^{b,\ast \ast}$}
\author[a]{Yiming Bie}
\author[b]{Wanjing Ma}

\cortext[cor1]{Corresponding author. E-mail: wqzheng@jlu.edu.cn}

\address[a]{School of Transportation, Jilin University,\\
5988 Renmin St., Changchun, Jilin 130022, China}
\address[b]{Key Laboratory of Road and Traffic Engineering of the Ministry of Education, College of Transportation, Tongji University, 4800 Cao'an Rd., Shanghai 201804, China}

\begin{abstract}
Cumulative culture enables humans to preserve, reuse, and extend knowledge and skills across experiences and generations. Inspired by this principle, we introduce \textit{SimSkill}, a self-evolving agent built around the Simulation of Urban MObility (SUMO) traffic simulator. SimSkill continually identifies capability gaps, generates and solves environment-grounded tasks, verifies solutions through an action--critic loop, and consolidates experience into episodic, procedural, and semantic memory. Through autonomous exploration, it builds a library of reusable skills and knowledge spanning major stages of the traffic-simulation workflow. We evaluate SimSkill on two held-out benchmarks across three backbone LLMs, with each result independently verified. It improves verified success by up to 25 percentage points, and ablations show complementary contributions from procedural and semantic memory. Its benefits remain backbone- and budget-dependent, as memory does not improve every model or uniformly reduce inference cost. More broadly, SimSkill illustrates a natural-language-centered design paradigm for LLM-based agent systems. Its high-level control logic, operating principles, and accumulated knowledge are expressed in natural language, while an LLM integrates them with executable tools and code to realize precise and reproducible execution. All code and experimental data are publicly available at \url{https://github.com/qiliuchn/SimSkill-V1}.
\end{abstract}

\end{frontmatter}

\section{Introduction}
\label{sec:introduction}

The long-term value of a system built on a large language model (LLM) depends not only on how well it reasons and generalizes when it has no prior experience, but also on whether it can turn its own interactions into durable, reusable competence. LLM-based autonomous agents combine language-model reasoning with planning, tool use, environmental feedback, and memory~\citep{wang2024survey,yao2022react,zhang2025survey}. Most deployed agents, however, gain little from use: those without persistent memory begin each session from an empty context, while those with it accumulate retrievable records rather than reusable competence. In both cases, the procedures, diagnoses, and domain facts developed while solving one task must be rediscovered for the next. The challenge is therefore to build agents that accumulate experience and turn it into capability that persists across tasks.

Traffic simulation is a demanding instance of this problem: infrastructure, operations, and control policies are evaluated in simulation before deployment~\citep{chen2024data}, and Simulation of Urban MObility (SUMO) is among the most widely used open-source platforms for the purpose~\citep{behrisch2011sumo}. Competent use nonetheless demands both transportation knowledge and software expertise, a substantial barrier for new users~\citep{ejercito2017traffic,haddouch2018modeling}. Recent LLM-based systems such as ChatSUMO~\citep{li2024chatsumo}, SUMO-MCP~\citep{ye2025sumo}, and ChatSUMO-Agent~\citep{li2026chatsumoagent} lower that barrier by mapping natural-language requests onto scripts and tools prepared in advance. They can devise new workflows at run time. However, the procedures and knowledge during execution are generally not consolidated into persistent capabilities for later tasks, so successful use does not necessarily expand the system's reusable repertorie. The task space is too diverse for a fixed repertoire to reach far beyond the cases its designers foresaw.

Retaining and reusing capability is the concern of lifelong or continual learning, in which an agent acquires and integrates new knowledge while preserving what it has already learned~\citep{zheng2026lifelong}. The pioneering Voyager system~\citep{wang2023voyager} and subsequent studies~\citep{zhu2023ghost,wang2024jarvis} showed that an LLM agent can pursue this objective through open-ended exploration in environments such as Minecraft. Three questions remain open: how to generate informative experience through self-directed exploration, how to distill that experience into transferable capability, and how to organize what has accumulated so that it remains usable as the collection grows.

We present \textit{SimSkill}, a self-evolving LLM agent that addresses these questions in SUMO. SimSkill treats SUMO as an environment to be explored: the agent proposes tasks that extend its current competence, attempts and evaluates them against simulation output, and consolidates the outcome into persistent memory. That memory is separated into three complementary forms: episodic records preserve complete attempts and their evidence, procedural skills pair adaptable natural-language instructions with executable resources, and semantic pages hold structured, cross-linked domain knowledge. Experience from individual tasks thereby becomes procedural and semantic artifacts that later tasks retrieve, compose, and revise. The main contributions of this work are as follows:
\begin{itemize}
    \item We formulate simulator mastery as a lifelong learning problem, and introduce SimSkill, a framework that closes the loop among autonomous curriculum generation, environment-grounded task execution, and experience consolidation.

    \item We develop a tripartite memory architecture (episodic, procedural, semantic) together with its complete lifecycle: bounded retrieval, ingestion, merging, and linting keep the stored artifacts reusable as the collection grows. In approximately 80 hours of autonomous operation over five days, SimSkill accumulated 150 procedural skills and 153 semantic-memory pages spanning the major stages of traffic-simulation practice. The resulting artifacts are inspectable, editable, composable, and transferable across LLM backbones and agent frameworks.

    \item We evaluate SimSkill on two held-out 40-task benchmarks with three LLM backbones, independent verification, and five-condition ablations. It improves verified success by up to 25 percentage points, and ablations show complementary contributions from procedural and semantic memory. Its benefits remain backbone- and budget-dependent: memory does not improve every model or uniformly reduce inference cost.
\end{itemize}

Although instantiated in SUMO, SimSkill illustrates a more general design principle for LLM-based agent systems. Its high-level control logic is specified through natural-language instructions rather than hard-coded workflows, so that language acts as the connective medium through which executable components are selected, adapted, and composed. This division of labor---language to preserve and recombine capability, code and tools to execute it precisely and reproducibly---is akin to the role language plays in human culture, where it both carries what has been learned and organizes how work is planned and how tools are combined. SimSkill can be extended to other executable environments that supply tools, observable outcomes, and reusable structure.

The remainder of this paper is organized as follows. Section~\ref{sec:related_work} reviews related work. Section~\ref{sec:method} presents the conceptual foundations, memory architecture, learning process, and SUMO instantiation of SimSkill. Section~\ref{sec:case_studies} traces representative learning and memory-maintenance episodes. Section~\ref{sec:experiments} presents the experiments and results, and Section~\ref{sec:conclusion} concludes the paper. The appendices provide the complete inventory of accumulated skills and knowledge pages, supplementary traces from the learning process, and representative benchmark executions showing how procedural and semantic memory are used during inference.

\section{Related Work}
\label{sec:related_work}

SimSkill lies at the intersection of four lines of research: self-improvement of language models, experience-driven lifelong learning, long-term memory for LLM agents, and LLM-based traffic simulation.

\subsection{Self-Improvement and Self-Evolution of Large Language Models}
Research on self-evolving LLMs asks how a model can generate the experience and feedback needed for its own improvement~\citep{tao2024survey}. One line of research constructs training data with reduced human supervision, bootstrapping instruction data from a model's own generations~\citep{wang2023self}, fine-tuning on high-confidence self-generated rationales~\citep{huang2023large}, or filtering model-generated samples by reward~\citep{gulcehre2023reinforced}. Recursive Introspection~\citep{qu2024recursive} extends the same principle to self-correction, fine-tuning a model to revise its own unsuccessful attempts rather than relying on prompting alone. All convert self-generated output into parameter updates.

A second line improves behavior through interaction and reflection rather than retraining: autotelic agents set their own goals and acquire competence by exploring under feedback~\citep{colas2023augmenting}, Reflexion~\citep{shinn2023reflexion} turns task feedback into verbal reflections that condition later attempts, and AppAgent~\citep{zhang2025appagent} explores an environment under feedback to build a reusable knowledge base, all without weight updates. Two ideas carry into SimSkill: experience can be actively generated rather than passively supplied, and both trials and reflection on them supply signal that guides future generation. Rather than a single self-improvement pass ending in updated weights or raw experience, SimSkill accumulates and maintains skills and knowledge in external procedural and semantic memory, leaving the backbone model untouched.

\subsection{Experience-Driven Lifelong Learning Agents}
Lifelong LLM agents acquire, retain, and transfer capability across an extended sequence of interactions~\citep{zheng2026lifelong}. Voyager~\citep{wang2023voyager} is the seminal example, coupling an automatic curriculum, iterative improvement from execution feedback, and an ever-growing library of executable Minecraft skills. Related agents add text-based or multimodal memory for long-horizon planning~\citep{zhu2023ghost,wang2024jarvis}, and Lifelong Robot Library Learning~\citep{tziafas2024lifelong} carries the principle into embodied manipulation.

More recent work treats skills as maintainable artifacts: AutoSkill~\citep{yang2026autoskill} abstracts reusable skills from interaction traces, SkillOpt~\citep{yang2026skillopt} keeps a skill edit only when it improves held-out performance, and LifelongAgentBench~\citep{zheng2025lifelongagentbench} finds raw experience replay limited by irrelevant content and context-window pressure. SimSkill introduces new memory representations and management mechanisms, and carries the paradigm into traffic simulation.

\subsection{Long-Term Memory for LLM Agents}
Long-term memory lets an agent use information held outside both the model parameters and the current context window~\citep{zhang2025survey}. Retrieval-augmented generation (RAG)~\citep{lewis2020retrieval} supplies external knowledge at inference time; successive systems store conversations, observations, and reflections for later use~\citep{zhong2024memorybank,park2023generative,liu2026gatsim}, let the model manage its own external memory~\citep{packer2023memgpt}, and consolidate, enrich, and link records as the store grows~\citep{chhikara2025mem0,salama2025meminsight,xu2026mem}.

A parallel engineering literature supplies artifact-level representations: Agent Skills package instructions, references, and executable resources as discoverable filesystem artifacts~\citep{anthropic2026skills}, while the LLM Wiki pattern~\citep{karpathy2026wiki}, formalized as the Open Knowledge Format~\citep{mcveety2026okf}, represents curated knowledge as a directory of interlinked Markdown pages with structured front matter. SimSkill combines these ideas into a tripartite memory architecture with memory management techniques that support the execution of complex tasks.

\subsection{Large Language Model Agents for Traffic Simulation}
LLMs have entered transportation research as natural-language interfaces to traffic data, models, and analysis packages~\citep{zhang2024trafficgpt,da2024openti}, as reasoning engines for traffic-signal control~\citep{lai2025llmlight,wang2024llm}, and as generative travelers inside mobility simulations~\citep{liu2025generative,liu2026gatsim}. ChatSUMO~\citep{li2024chatsumo} maps user requests onto parameters for prepared SUMO scripts. SUMO-MCP~\citep{ye2025sumo} exposes SUMO's preprocessing, execution, optimization, and analysis utilities through the Model Context Protocol (MCP), letting an agent discover and chain tools at run time. AgentSUMO~\citep{jeong2025agentsumo} turns incomplete policy objectives into executable simulation plans, TrafficSimAgent~\citep{du2025trafficsimagent} pairs high-level agents that orchestrate the MCP tool workflow with low-level agents that control individual traffic elements, and ChatSUMO-Agent~\citep{li2026chatsumoagent} couples planning and multi-tool execution with simulation feedback to refine a solution within a single study. Across these systems, the primary goal is ease of use: users specify tasks in natural language, and agents execute them using predefined workflows and tools. SimSkill addresses a complementary question: how an agent can expand this repertoire through experience.

\section{Method}
\label{sec:method}

\subsection{Conceptual Foundations and Framework Overview}
\label{sec:method_overview}

SimSkill rests on four premises about how an LLM agent can turn interaction with an executable environment into cumulative competence.

\paragraph{1. Verification asymmetry enables self-improvement.}
For many engineering tasks, constructing a valid solution is open-ended and difficult, whereas evaluating a candidate decomposes into simpler checks. A traffic-simulation workflow, for example, can be assessed for syntactic validity, successful execution, compliance with task constraints, nonempty outputs, reproducibility, and consistency between reported metrics and generated artifacts. This asymmetry is what makes self-improvement possible: an agent can iteratively generate, test, diagnose, and revise candidate solutions against verifiable feedback.

\paragraph{2. Skill composition compounds capability.}
SimSkill draws on two ideas from reinforcement learning. The first is that competence grows through interaction, trial and error, and feedback~\citep{sutton1998reinforcement}, which multi-objective formulations extend to agents optimizing several competing signals at once. The second is skill-based reinforcement learning's emphasis on reusable and composable behaviors~\citep{sutton1999options, pertsch2021accelerating}. SimSkill practices both: it attempts diverse tasks by trial and error, and the skills and knowledge it acquires compose into progressively more complex capability. The connection is conceptual rather than algorithmic. SimSkill's feedback is more general than a scalar reward---textual completion judgments and error reports---and what it updates is external memory rather than model parameters.

\paragraph{3. General mechanisms, not hand-crafted structure, produce broad competence.}
Sutton's Bitter Lesson~\citep{sutton2019bitterlesson} holds that general methods that scale with computation eventually outperform systems built on hand-crafted domain structure. SimSkill follows this principle with a small set of general operations---task proposal, memory retrieval, environment interaction, criticism, consolidation, and maintenance---rather than a dedicated chain-of-thought template, planner, or hard-coded workflow for each traffic-simulation problem. Domain knowledge is accumulated as experience rather than embedded in the system architecture. SimSkill's curriculum favors novelty, diversity, practical value, gap coverage, and progressively increasing difficulty. These criteria guard against a system-level analogue of reward hacking, in which the agent collects repeated positive verdicts on a narrow family of tasks without expanding its competence. This broad objective distinguishes SimSkill's self-evolution from a deep-research process optimized only for the specific request.

\paragraph{4. Natural language is a better substrate for cumulative memory than parameter updates.}
Parametric knowledge is difficult to inspect, revise locally, attribute to evidence, or transfer between model families. Natural-language instructions and knowledge pages, by contrast, can be read, critiqued, combined, and shared by humans and heterogeneous LLMs. Prompt- and skill-based engineering suggests that carefully preserved instructions can constitute reusable capability~\citep{anthropic2026skills,multica2026karpathyskills}, much as humans accumulate skills and knowledge.

\begin{figure}
    \centering
    \includegraphics[width=\textwidth]{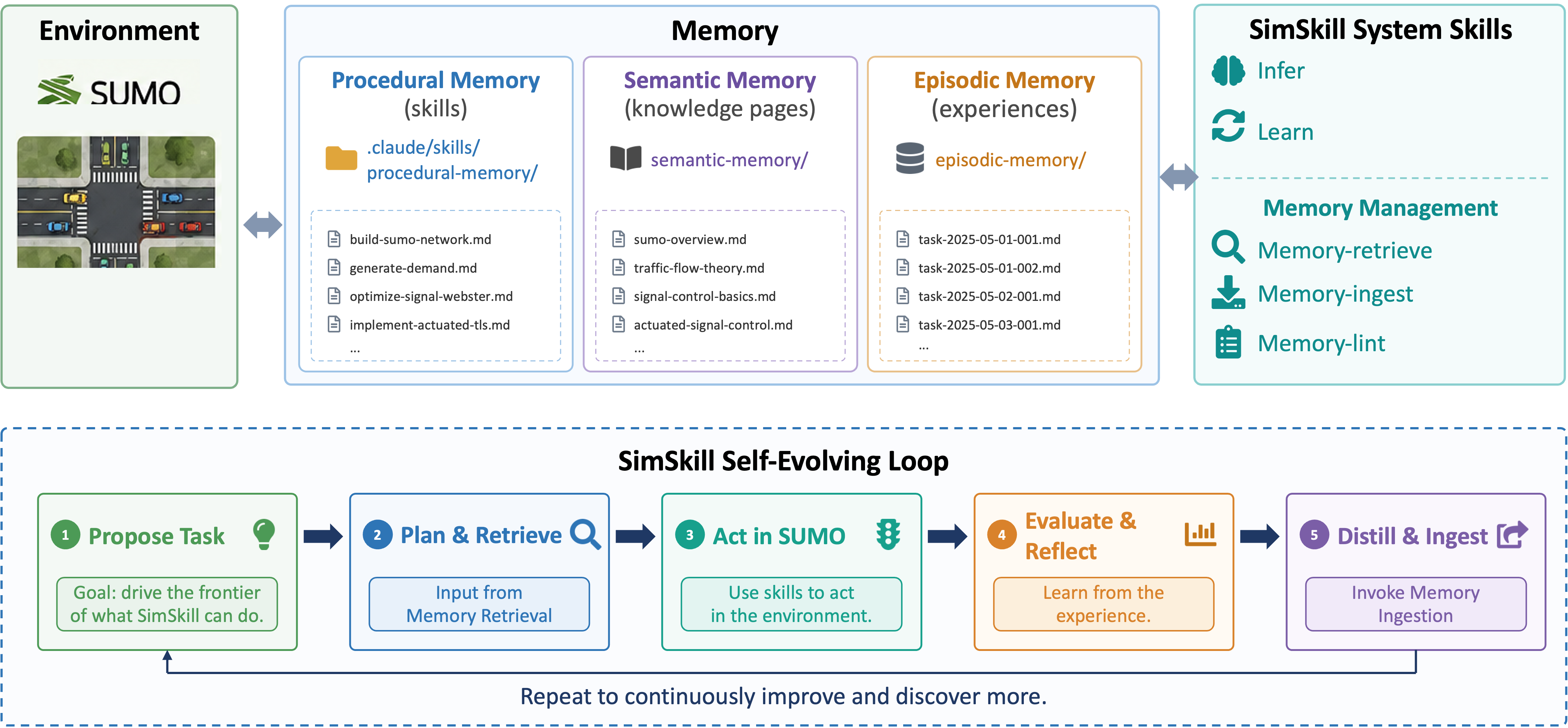}
    \caption{Architecture of SimSkill. A language-model runtime coordinates interaction with SUMO and three complementary external memory stores. SimSkill system skills implement inference, autonomous learning, memory retrieval, ingestion, and linting, while the lower loop summarizes continual task proposal, memory-guided action, evaluation, and consolidation.}
    \label{fig:simskill_framework}
\end{figure}

\paragraph{SimSkill Architecture.}
These four premises shape the architecture shown in Figure~\ref{fig:simskill_framework}. At learning iteration $t$, the persistent state of SimSkill is
\begin{equation}
    \mathcal{M}_{t} = \left(\mathcal{E}_{t}, \mathcal{P}_{t}, \mathcal{S}_{t}\right),
    \label{eq:memory_state}
\end{equation}
where $\mathcal{E}_{t}$, $\mathcal{P}_{t}$, and $\mathcal{S}_{t}$ denote episodic, procedural, and semantic memory, respectively. The system evolves by adding, revising, linking, validating, and consolidating these explicit artifacts.

Claude Code provides the filesystem, tool-use, and sub-agent runtime; the LLM backend can be replaced by any compatible model without altering the stored memory. Five core natural-language system skills specify the control logic: \texttt{learn}, \texttt{infer}, \texttt{memory-retrieve}, \texttt{memory-ingest}, and \texttt{memory-lint}. Three role-specialized agents---\texttt{curriculum-agent}, \texttt{action-agent}, and \texttt{critic-agent}---propose tasks, execute them, and independently evaluate the resulting evidence. The loop is simple: propose a task, retrieve relevant memory, act in the environment, evaluate the result, distill reusable outcomes, and lint the memory; the next iteration begins from the updated memory state. The exploration mechanism is inspired by Voyager~\citep{wang2023voyager}, while the richer memory structure lets SimSkill retain not only executable behavior but also declarative knowledge and the evidence supporting it.

\subsection{Tripartite Memory and Experience Consolidation}
\label{sec:method_memory}

Inspired by cognitive theories of memory~\citep{tulving2000oxford,squire2004memory}, we distinguish among working, episodic, procedural, and semantic memory. Working memory---the active task context---is managed by Claude Code; it is not part of the persistent state in Equation~\ref{eq:memory_state} and falls outside the scope of this work. The purpose of SimSkill's persistent memory is not merely to retain more text, but to preserve different products of learning at appropriate levels of abstraction. Episodic memory records \emph{what happened}, procedural memory records \emph{how to act}, and semantic memory records \emph{what is known}. Procedural and semantic memory can be viewed as two complementary compressions of episodic experience: the former abstracts reusable patterns of action, whereas the latter abstracts regularities about the environment.

\paragraph{Episodic memory.}
Every attempted task produces a timestamped episode in \texttt{episodic-memory/}. An episode contains the original task, its success status, the memory items used, and the complete sequence of action--critic attempts. For each attempt, SimSkill preserves the action agent's report, the critic's evidence and verdict, and any scripts created or executed. It also stores the final deliverables and a summary recording the method, reproducible commands, measured results, and the transition between successive attempts. Failed attempts are retained rather than overwritten by the final solution. Episodic memory therefore serves as an auditable evidence layer: it records not only the answer reached, but also how claims were tested and why revisions were made.

\paragraph{Procedural memory.}
Procedural memory is stored as Claude Code skills~\citep{anthropic2026skills}. Each skill is a directory containing a required \texttt{SKILL.md} file and optional \texttt{scripts/}, \texttt{references/}, and \texttt{assets/} directories. YAML front matter provides a unique name and a retrieval-oriented description, while the Markdown body describes when and how the procedure applies, including assumptions, validation checks, execution process, and known failure modes. Executable scripts capture operations that must be repeated exactly. A skill may invoke simpler skills or link to relevant knowledge pages, so that complex capabilities are built compositionally. Unlike a library in which each skill is a single fixed function~\citep{wang2023voyager}, this representation combines flexible natural-language strategy with reproducible executable components.

The representation has two further advantages. First, one format spans the range from flexible interaction to exact execution: natural-language instructions express preferences, heuristics, explanation conventions, and context-dependent judgments, whereas bundled scripts implement operations that must run identically each time. A single skill may combine both, letting the LLM adapt the procedure to a new situation while delegating its deterministic steps to code. Second, externalized skills make acquired capabilities transparent, editable, and attributable: users and developers can inspect the rules and scripts, revise an incorrect assumption locally, follow links to supporting knowledge and experience, and identify which retrieved skills influenced a later solution, since skill use is recorded in the action report and in episodic memory.

Consequently, a completed interaction need not remain an ephemeral transcript or an opaque change in model behavior. SimSkill distills its reusable procedural content into a persistent artifact that can be retrieved across sessions, refined when new evidence arrives, composed with other skills, and shared without modifying the backbone model. This explicitness is what makes competence cumulative efficient and easy to implement.

Listing~\ref{lst:procedural_memory_format} gives the file-level contract. The description is deliberately operational: it is the first-stage retrieval key and must state both what the skill does and when it should be selected. Detailed references and deterministic operations are kept out of the instruction body so that they are loaded or executed only when needed.

\begin{lstlisting}[basicstyle=\small\ttfamily,
                    caption={Directory structure and core content of a procedural-memory skill.},
                   label={lst:procedural_memory_format}]
(a) Skill directory structure:
your-skill-name/
|-- SKILL.md                 # required: metadata and instructions
|-- scripts/                 # optional: executable utilities
|-- references/              # optional: supporting documentation
|   |-- api-guide.md
|   `-- examples/
`-- assets/                  # optional: templates and resources
    `-- report-template.md

(b) SKILL.md metadata and instructions:
---
name: your-skill-name
description: Concisely state what the skill does and when to use it.
---

# <Skill title>
Natural-language instructions specifying procedures,
assumptions, validation criteria, known failure modes,
and links to related skills and knowledge pages.
\end{lstlisting}

\paragraph{Semantic memory.}
Semantic memory is a persistent, LLM-maintained knowledge base inspired by the LLM Wiki pattern~\citep{karpathy2026wiki} and represented in a form compatible with Open Knowledge Format~\citep{mcveety2026okf}. Each concept is stored as one Markdown page in \texttt{semantic-memory/}. Its front matter contains a summary, retrieval keywords, creation and update times, provenance sources, related pages, and related skills. Wiki links connect concepts into a graph, and citations point either to external sources or to source material retained in \texttt{raw-materials/}. A compact index exposes the summary and keywords of every page, so the agent can search the knowledge base without loading page bodies. The result is synthesized, organized, and revisable knowledge rather than a fixed collection of raw chunks, as in conventional RAG.

\begin{figure}
    \centering
    \includegraphics[width=0.94\textwidth]{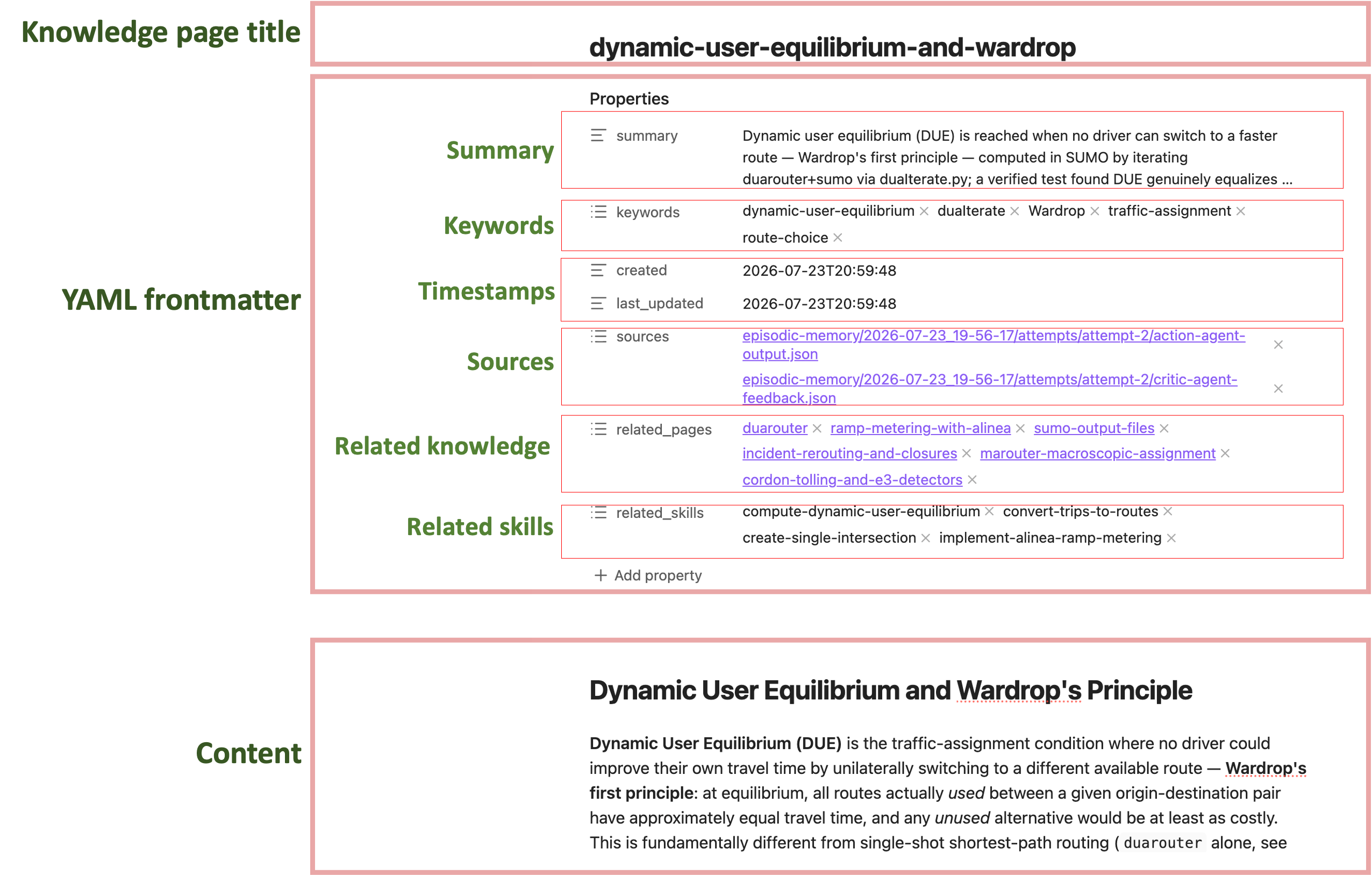}
    \caption{An example semantic-memory knowledge page. The YAML front matter contains the retrieval summary, keywords, timestamps, provenance sources, related knowledge pages, and related procedural skills; the Markdown body stores the synthesized declarative content. The cross-links to related pages and skills, organize memory into a navigable graph: they extend retrieval beyond the initially matched page and route updates to the items a new lesson affects---the property that distinguishes this store from the flat chunk collections of conventional RAG.}
    \label{fig:knowledge_page_format}
\end{figure}

Listing~\ref{lst:semantic_memory_format} shows knowledge page representation. The \texttt{summary} and \texttt{keywords} fields are copied into \texttt{semantic-memory/index.md} for first-stage retrieval; \texttt{sources} preserve provenance; and \texttt{related\_pages} and \texttt{related\_skills} connect the page to the other pages or procedures that use it. These links can serve second-stage retrieval and make memory maintenance more efficient. Figure~\ref{fig:knowledge_page_format} shows how the schema appears in an Obsidian-rendered knowledge page.

\begin{lstlisting}[basicstyle=\small\ttfamily,
                    caption={Semantic memory knowledge page (\texttt{semantic-memory/<knowledge-page-title>.md}) format.},
                   label={lst:semantic_memory_format}]
---
summary: One or two sentences describing the concept.
keywords: [keyword-1, keyword-2]
created: YYYY-MM-DDThh:mm:ss
last_updated: YYYY-MM-DDThh:mm:ss
sources:
  - "[[raw-materials/source-file.md]]"
  - https://example.com/source
related_pages: ["[[related-concept-1]]", "[[related-concept-2]]"]
related_skills: [related-skill-1, related-skill-2]
---

# <Knowledge Page Title>
Synthesized facts, explanations, qualifications, and links to 
related concepts, source materials, and procedural skills.
\end{lstlisting}

\paragraph{Retrieval.}
Given a task $q$, memory retrieval first matches $q$ against skill descriptions and the summaries and keywords in the semantic index. It returns a bounded set of candidate skills and pages
\begin{equation}
    \mathcal{R}(q) = \operatorname{TopK}\!\left(q;\mathcal{P}_{t}\cup\mathcal{S}_{t}\right),
    \qquad \left|\mathcal{R}(q)\right| \leq N.
    \label{eq:memory_retrieval}
\end{equation}
Selection and loading are separate steps: $\mathcal{R}(q)$ is determined from metadata alone, whereas the full content of a candidate is read into context only when the task requires it. This lazy-loading policy limits context growth while preserving access to a much larger external store. Raw episodes are not routinely replayed during task execution; they remain available for audit and for curriculum decisions---especially when a previously failed task may now be solvable---while their transferable content is consolidated into procedural or semantic memory.

\paragraph{Consolidation.}
After an episode $e_t$ has been completed and evaluated, the memory-ingestion process applies an LLM-based consolidation operator
\begin{equation}
    \left(\mathcal{P}_{t+1},\mathcal{S}_{t+1}\right)
    = \mathcal{I}\!\left(\mathcal{P}_{t},\mathcal{S}_{t},e_t\right).
    \label{eq:memory_consolidation}
\end{equation}
The operator first asks whether the episode contains anything novel and reusable. A procedural lesson may create a new skill, extend an existing skill to a broader case, or correct a previously discovered defect; a declarative lesson may create or revise a knowledge page. Both may occur for the same episode, and ingestion is skipped when no reusable contribution is found. Before creating an artifact, the agent searches for a semantically similar item and updates it where appropriate, favoring cumulative refinement over near-duplicate proliferation. New and revised pages are cross-linked to related pages and skills, the semantic index is synchronized, and every change is appended to a shared log. Consolidation thus compresses experience along two axes: \emph{what appears to be true} and \emph{how a class of tasks can be performed}. Listing~\ref{lst:memory_system_skills} presents the operational core of the two system skills that connect active reasoning to long-term memory. It is a condensed transcription of their natural-language instructions.

\begin{lstlisting}[basicstyle=\small\ttfamily,
                    caption={Memory retrieval and ingestion system skills.},
                   label={lst:memory_system_skills}]
(a) SYSTEM SKILL: memory-retrieve(task)
1. Match the task against description front matter in
   procedural memory.
2. Match the task against summaries and keywords in the
   semantic-memory index.
3. If more than ten items match, retain the ten most relevant
   skills and knowledge pages in total.
4. Return their entries as context; load full contents lazily
   only when the action agent needs them.

(b) SYSTEM SKILL: memory-ingest(episode)
1. Read the task, action-agent report, and critic feedback.
2. Identify novel and reusable procedural or semantic lessons;
   stop if the episode adds none.
3. For a procedural lesson, update a similar skill or create a
   new skill, preferring composition over reimplementation.
4. For a semantic lesson, update a similar page or create a new
   knowledge page.
5. Add page-to-page and page-to-skill links, and synchronize the
   semantic-memory index.
6. Record every accepted change in the shared memory log.
\end{lstlisting}

\paragraph{Memory maintenance.}
The \textit{stability--plasticity problem} is a fundamental dilemma in continual, or lifelong, learning: how a system can acquire new knowledge without destroying what it has already learned~\citep{wang2024comprehensive}. Explicit memory does not eliminate this dilemma, but converts it from an opaque problem of parameter updating into a tractable problem of artifact management. New evidence can be incorporated locally while existing content remains visible and can be tested for compatibility: a knowledge page is updated in place when new information arrives, and a skill is revised when its interface must change to cover broader cases. Because no model parameters are updated, SimSkill also sidesteps parameter-level catastrophic forgetting~\citep{kirkpatrick2017overcoming}.

Maintenance is performed periodically by \texttt{memory-lint}, which compares newly changed items against the full collection, merges genuine duplicates, repairs broken or drifted references, removes fully superseded artifacts, and synchronizes the semantic index. Incremental linting is triggered after a configurable number of changes (ten by default); full linting checks the entire collection and can also be invoked manually. A shared append-only log lets the process determine its incremental scope without loading the growing history into the LLM context. Externally contributed memory---for example, from a public repository---is processed by \texttt{memory-merge}: new skills are executed on representative tasks before acceptance, and updates are rejected when compatibility with local dependents cannot be established. The current implementation relies on consolidation, supersession, and merging rather than age-based decay; automatic forgetting by access frequency is left for future work. Listing~\ref{lst:memory_lint_skill} condenses the workflow of \texttt{memory-lint}.

\begin{lstlisting}[basicstyle=\small\ttfamily,
                    caption={Memory maintenanc system skill.},
                   label={lst:memory_lint_skill}]
SYSTEM SKILL: memory-lint(mode=incremental)
1. Query the shared log for changed-item names and count.
2. Determine scope:
   a. incremental: stop below the change threshold; otherwise
      inspect changed items against the full memory collection;
   b. full: inspect every procedural skill and semantic page.
3. For procedural memory:
   a. merge only genuine duplicates;
   b. repair drifted names, paths, arguments, and interfaces;
   c. replace standalone procedures with verified compositions
      where appropriate;
   d. remove superseded skills and redirect their dependents.
4. For semantic memory:
   a. merge genuine duplicates and remove superseded pages;
   b. repair page links and related-skill references.
5. Synchronize the semantic index with the resulting pages.
6. Record every change, close the current lint interval, and
   open the next interval in the shared log.
7. Report merges, repairs, removals, and preserved distinctions.
\end{lstlisting}

\begin{figure}
    \centering
    \includegraphics[width=\textwidth]{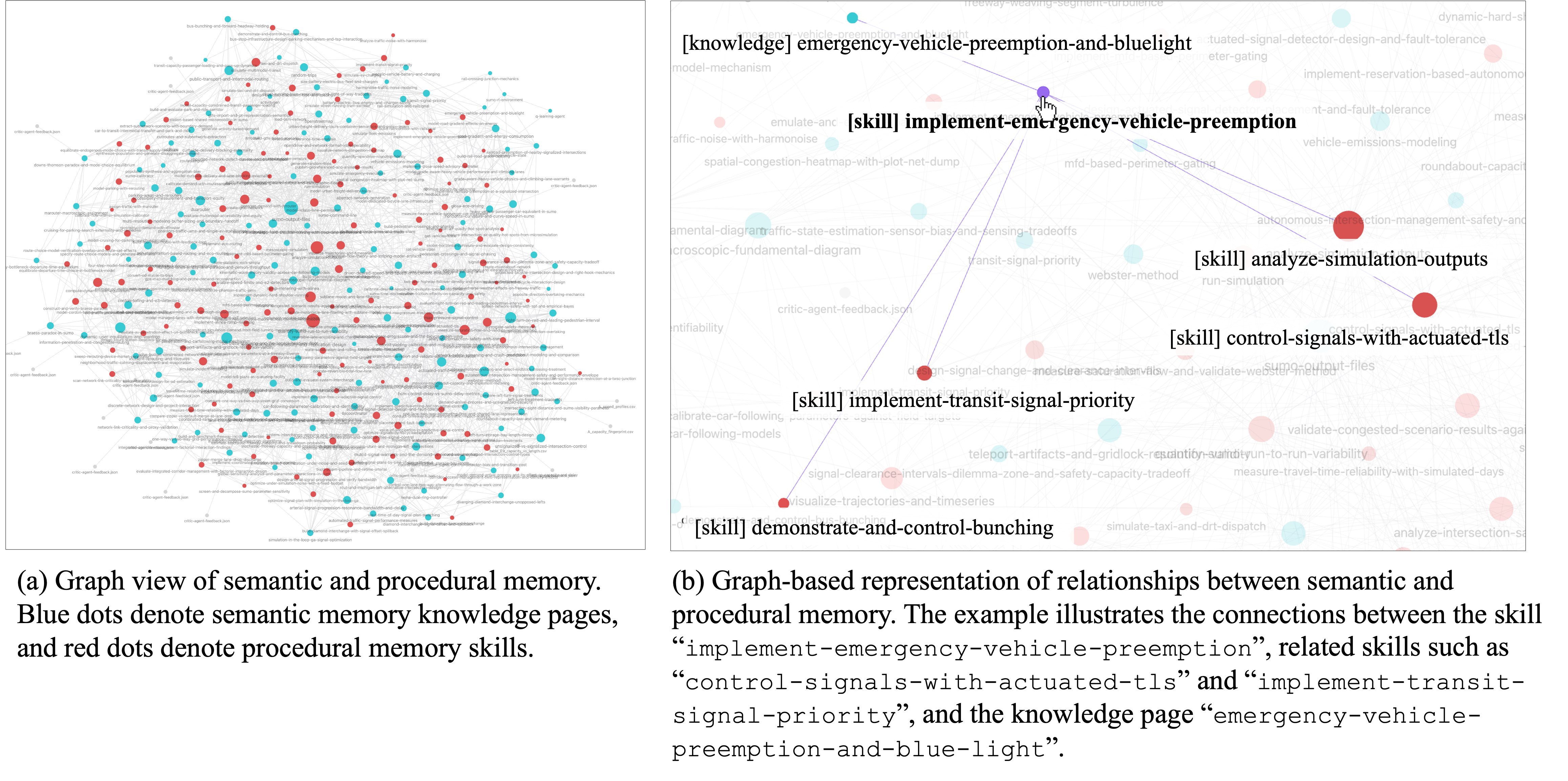}
    \caption{Graph view of accumulated procedural and semantic memory rendered by Obsidian. The left panel shows the collection-level network of skills, knowledge pages, and their explicit links; the right panel expands a local neighborhood around emergency-vehicle-preemption knowledge and its related procedural skills.}
    \label{fig:memory_graph}
\end{figure}

The explicit \texttt{related\_pages} and \texttt{related\_skills} fields make the accumulated structure inspectable as a graph. Figure~\ref{fig:memory_graph} shows both the full procedural--semantic memory network and a local neighborhood centered on skill \texttt{implement-emergency-vehicle-preemption} rendered by Obsidian.

\subsection{Self-Evolution}
\label{sec:method_learning}

\begin{figure}[H]
    \centering
    \includegraphics[width=0.94\textwidth]{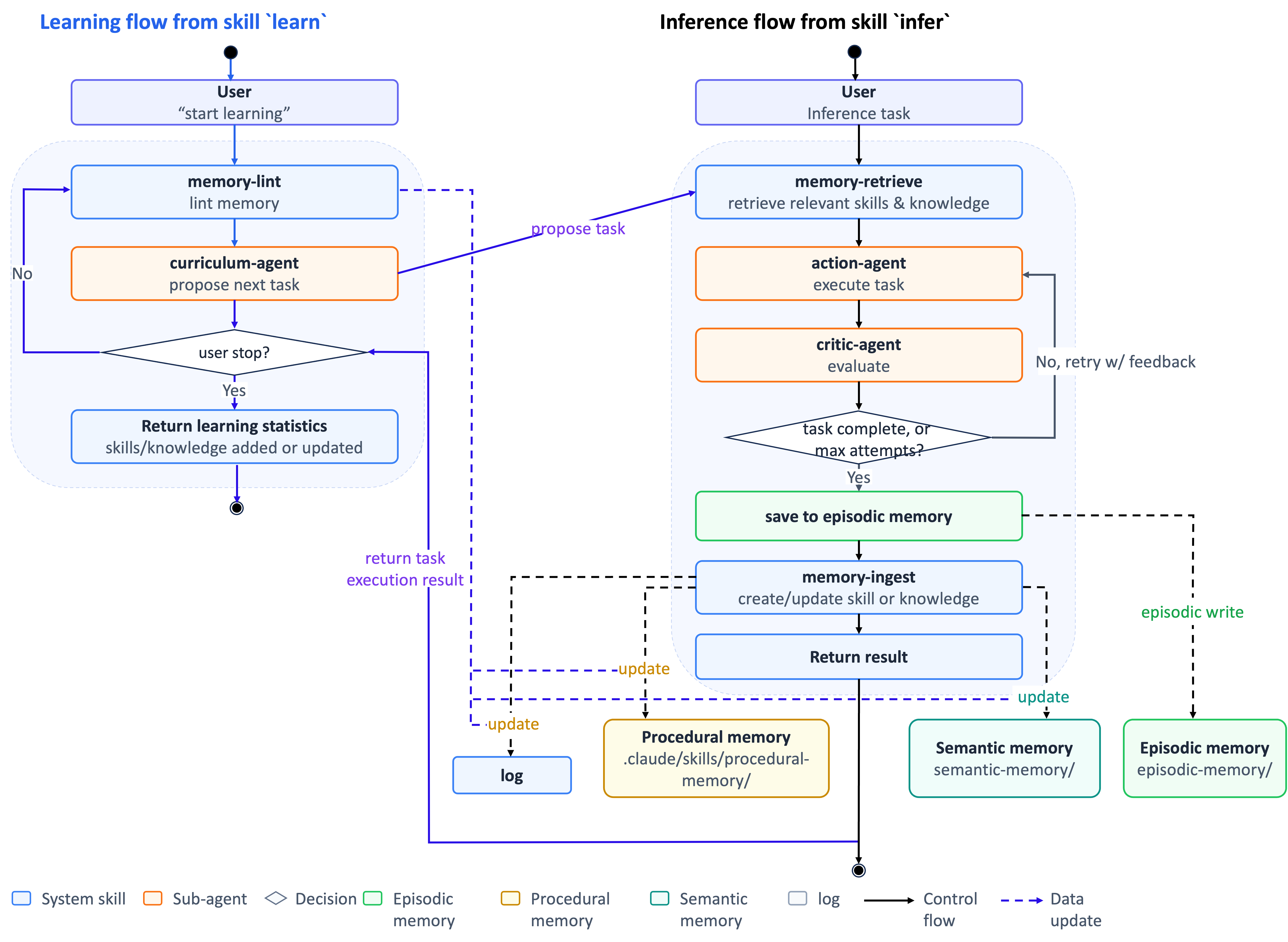}
    \caption{Detailed SimSkill learning, inference, and memory-management workflows. Autonomous learning (left) uses direct inference (right) as an inner process: the curriculum agent proposes a task, inference retrieves memory and iterates between execution and criticism, completed attempts are written to episodic memory, and reusable outcomes update procedural and semantic memory.}
    \label{fig:simskill_workflow}
\end{figure}

Autonomous learning is implemented as a continual curriculum--execution--consolidation loop, shown on the left of Figure~\ref{fig:simskill_workflow}. At the beginning of each iteration, SimSkill runs incremental memory maintenance and invokes the curriculum agent. Let $\mathcal{E}^{\mathrm{fail}}_t\subseteq\mathcal{E}_t$ denote prior unsuccessful episodes. The next task is proposed as
\begin{equation}
    q_t = \mathcal{C}\!\left(\mathcal{P}_t,\mathcal{S}_t,\mathcal{E}^{\mathrm{fail}}_t\right),
    \label{eq:curriculum}
\end{equation}
where $\mathcal{C}$ is realized by the curriculum agent's natural-language policy. The agent inspects current skill coverage ($\mathcal{P}_t$), the semantic index ($\mathcal{S}_t$), and prior failures ($\mathcal{E}^{\mathrm{fail}}_t$), identifies a capability gap, and explains why the proposed task is the appropriate next step.

Let $\mathcal{Q}$ denote a broad family of tasks and let $\operatorname{Perf}(q;\mathcal{M}_t)$ denote the performance of the agent on task $q$ with memory state $\mathcal{M}_t$. The conceptual lifelong-learning objective is
\begin{equation}
    J(\mathcal{M}_t)
    = \mathrm{E}_{q\sim\mathcal{Q}}
      \left[\operatorname{Perf}(q;\mathcal{M}_t)\right],
    \label{eq:broad_competence}
\end{equation}
rather than performance on one repeatedly optimized task~\citep{zheng2026lifelong}. SimSkill does not estimate this expectation during learning; it uses the objective to guide curriculum design. Here $\mathcal{Q}$ is taken to be traffic-simulation research and practical applications, and the proposed tasks aim to cover that space. Aligning the distribution of proposed tasks with the intended task space is essential, and requires careful design of the curriculum agent together with human-in-the-loop guidance.

A proposed task must be novel or a genuine extension of an existing capability, and achievable with current competence plus a reasonable increment of new work; across iterations the curriculum favors diversity, practical value, gap coverage, and progressively increasing difficulty. These criteria reduce the risk that the agent obtains repeated positive verdicts by exploiting a narrow family of tasks---a system-level analogue of reward hacking---while failing to expand its overall competence. 

The core control flow is itself represented as natural-language system skills. Listing~\ref{lst:learning_inference_skills} condenses the operative steps of \texttt{learn} and \texttt{infer}. The former is the persistent autonomous-learning loop; the latter is a complete task-solving transaction and is invoked both by \texttt{learn} and by direct user requests. The system skills delegate bounded responsibilities to three role-specialized sub-agents. Listing~\ref{lst:subagent_contracts} summarizes their input, task, and output contracts; implementation-specific prompt wording is omitted.

\begin{lstlisting}[basicstyle=\small\ttfamily,
                    caption={Core autonomous-learning and inference system skills.},
                   label={lst:learning_inference_skills}]
(a) SYSTEM SKILL: learn
1. Load memory-lint and infer.
2. Repeat:
   a. Run memory-lint in incremental mode.
   b. Ask curriculum-agent for the next novel task.
   c. Invoke infer on that task in normal mode.
3. Stop when the user requests; report memory changes.
4. Pause for guidance after ten iterations with no memory change.

(b) SYSTEM SKILL: infer(task, mode=normal)
1. Retrieve task-relevant procedural and semantic memory.
2. Initialize an empty critic-feedback history.
3. Repeat for at most three outer attempts:
   a. Invoke action-agent with the task, retrieved memory, and
      all accumulated critic feedback. The action agent may
      internally retry script execution up to five times.
   b. Invoke critic-agent to independently evaluate the result.
   c. Append the critic feedback to the feedback history.
   d. Stop if the task is complete; otherwise retry step 3a.
4. Save every outer attempt, verdict, script, and final output
   to episodic memory.
5. In normal mode, invoke memory-ingest; in test mode, skip it.
\end{lstlisting}

\begin{lstlisting}[basicstyle=\small\ttfamily,
                    caption={Role contracts of the three SimSkill sub-agents.},
                   label={lst:subagent_contracts}]
(a) AGENT: curriculum-agent
Input:  Procedural-memory coverage, semantic-memory index,
        and previously failed episodes.
Role:   Identify a capability gap and propose one novel,
        useful, achievable, and progressively harder task.
Output: Gap-based reasoning and a concrete task specification.

(b) AGENT: action-agent
Input:  Task, retrieved skills and knowledge pages, and any
        accumulated critic feedback.
Role:   Reuse available memory, create and execute required
        artifacts, inspect outputs, and repair minor execution
        errors for up to five attempts.
Output: Memories used, method, scripts, reproduction commands,
        measured results, and a concise retry history.

(c) AGENT: critic-agent
Input:  Task and the action agent's report and artifacts.
Role:   Independently check reproducibility, claims against
        evidence, silent failures, and coverage of task scope.
Output: Evidence, a completion verdict, and actionable feedback
        when revision is required.
\end{lstlisting}

Each proposed task is passed to \texttt{infer} process. The retriever supplies $\mathcal{R}(q_t)$, and the action agent constructs a solution using those memories together with native environment tools. It writes and executes the required scripts, inspects simulator output, and corrects minor execution errors internally. Its output is a structured report containing the memories used, created artifacts, exact reproduction commands, measured results, and any retries. Requiring execution and concrete outputs distinguishes an environment-grounded experience from an untested textual answer.

An independently prompted critic then evaluates the candidate. For outer attempt $j$, the interaction can be summarized as
\begin{equation}
    \begin{aligned}
        x_t^{(j)} &= \mathcal{A}\!\left(q_t,\mathcal{R}(q_t),f_t^{(<j)}\right), \\
        \left(v_t^{(j)},f_t^{(j)}\right) &= \mathcal{V}\!\left(q_t,x_t^{(j)}\right),
    \end{aligned}
    \label{eq:action_critic}
\end{equation}
where $x_t^{(j)}$ is the action agent's executed solution, $v_t^{(j)}$ is the completion verdict, and $f_t^{(j)}$ is evidence-based feedback. The critic checks reproducibility, correspondence between claims and artifacts, silent failures such as empty outputs or unrouted demand, and coverage of every requested subtask.

During development, the workflow frequently terminated at the action agent's report when that report described a successful execution. Because the critic was never invoked, unevaluated experience could reach the consolidation stage. A key orchestration rule therefore treats such a report as an intermediate result: control must return to \texttt{infer}, and explicit reminders to that effect were added to \texttt{infer} and to the action agent.

The action agent and \texttt{infer} maintain two nested retry loops at different abstraction levels. The \emph{inner execution loop} belongs to one invocation of the action agent. It addresses local implementation failures---for example, syntax errors, incorrect paths or arguments, simulator runtime errors, and malformed or empty output---by modifying and rerunning scripts until execution succeeds or the attempt limit is reached. The action agent then emits one structured report for that outer attempt, including a concise account of its failures and retries. The \emph{outer design loop} belongs to \texttt{infer}. After the action agent returns, the critic asks whether the executed solution actually satisfies the task. A negative verdict may expose a wrong modeling assumption, an omitted requirement, an invalid evaluation design, or a silent failure that successful execution alone did not reveal. \texttt{infer} then invokes a new action agent, up to a default limit of outer attempts.

Information crosses the loop boundary selectively. Every critic response $f_t^{(j)}$ is appended to the feedback history $f_t^{(<j+1)}$ and supplied to the next action-agent invocation, preventing previously identified design errors from being repeated. By contrast, the raw command-by-command debugging trajectory of the inner loop is not inserted into later outer contexts. Its structured report---including a concise failures-and-retries account---and the produced scripts and artifacts are retained as the record of that outer attempt. All outer action reports and critic verdicts are subsequently preserved in episodic memory. This separation keeps routine debugging from consuming the design-revision context while retaining the feedback needed for cumulative correction.

When the critic accepts the result or the attempt limit is reached, SimSkill writes the full episode to $\mathcal{E}_{t+1}$ and invokes the consolidation operator in Equation~\ref{eq:memory_consolidation}. Ingestion is based on reusable evidence rather than success alone. A failed task may still update a skill with a newly discovered limitation or create semantic knowledge about an invalid modeling assumption. Conversely, a successful task need not change memory if it merely repeats known procedures. The next curriculum iteration therefore operates on the capabilities and gaps revealed by the previous one. Learning continues until the user stops it; as a safeguard against unproductive cycling, the current implementation pauses after ten consecutive iterations that produce no procedural or semantic change.

\paragraph{Preventing context explosion.}
SimSkill's external memory grows throughout its lifetime, and a single task can generate a long execution trace. The framework treats the context window as a bounded working resource and makes use of three mechanisms to limit active-context consumption.

First, \textit{memory retrieval is metadata-first and lazy}. The retrieval skill initially searches only skill descriptions and the summaries and keywords in the semantic-memory index, retains at most ten candidates across the two memory types, and loads complete skill definitions, scripts, references, or knowledge pages only when they are needed. Raw episodic records are likewise not replayed by default, because their transferable content is consolidated into procedural and semantic memory during ingestion. The active context therefore does not grow in proportion to the size of persistent memory.

Second, \textit{routine operations are delegated to lightweight script tools}. For example, the system's \texttt{log} skill updates the append-only \texttt{log.md} through log manager script API instead of placing the entire, continually growing log in the active context. The active context thus does not grow in proportion to the length of the memory change history, most of which is irrelevant to the current step.

Third, \textit{role-specialized sub-agents isolate transient details in separate contexts}. The curriculum, action, and critic agents receive only the information required by their roles and return fixed-schema reports to the orchestration loop. In particular, accumulated critic feedback is passed to the action agent across outer inference attempts so that major design mistakes are not repeated, whereas command-level debugging traces from the action agent's inner script-repair loop remain local and are not copied into subsequent attempts. The active context therefore grows with the number of outer attempts rather than with the length of the execution trace.

\subsection{Learning-Time and User-Directed Task Execution}
\label{sec:method_inference}

In the current implementation, the same \texttt{infer} skill serves both learning-time task execution and user-directed inference. The autonomous \texttt{learn} process invokes it after the curriculum agent proposes a task, whereas a user request enters it directly, bypassing curriculum generation and routine linting. In both cases, SimSkill retrieves relevant procedural and semantic memory and applies the action--critic protocol shown on the right of Figure~\ref{fig:simskill_workflow}. The \texttt{infer} skill also supports two memory-write policies. Normal mode records the episode and considers it for procedural or semantic ingestion, allowing user interactions as well as autonomous tasks to improve future performance. Test mode retains the episode for reproducibility but disables ingestion, leaving procedural and semantic memory frozen for controlled evaluation.

Learning and inference sharing one execution policy is an implementation choice rather than an architectural requirement. Learning-time execution benefits from stringent criticism, retries, and artifact checks because an undetected error may be consolidated into memory and influence many future tasks. User-facing inference may instead prioritize latency or cost; a deployment could omit the critic verdict, reduce retry limits, or disable write-back, accepting a corresponding reduction in assurance. Conversely, high-stakes applications may require stronger verification than autonomous exploration. Listing~\ref{lst:fast_user_inference} illustrates an inference process for time-constrained user requests.

\begin{lstlisting}[basicstyle=\small\ttfamily,
                    caption={Illustrative fast inference skill for time-constrained user requests.},
                   label={lst:fast_user_inference}]
SYSTEM SKILL: infer-fast(task)
1. Load memory-retrieve.
2. Retrieve task-relevant procedural and semantic memory.
3. Invoke action-agent in the foreground with the task and
   retrieved memory.
4. Let action-agent execute the required artifacts and repair
   minor script errors within its local execution loop.
5. If execution succeeds, return its method, results,
   reproduction commands, and artifacts directly to the user.
6. If the execution limit is reached without success, return
   an incomplete status and a concise diagnostic summary.
\end{lstlisting}

\subsection{Instantiation in the SUMO Environment}
\label{sec:method_sumo}

SimSkill instantiates the general framework in SUMO~\citep{behrisch2011sumo,sumo2026}. The environment exposes diverse, compositional operations: constructing and validating road networks, synthesizing and routing demand, configuring signal controllers, executing microscopic or mesoscopic simulations, interacting with vehicles and infrastructure at runtime, and analyzing quantitative outputs. These operations supply both an open-ended capability space and concrete feedback, making SUMO suitable for studying cumulative agent learning.

Domain specialization enters at three points. First, a project-level \texttt{CLAUDE.md} file supplies persistent orientation: the target environment, where each memory store resides, and the format its artifacts follow. Second, the curriculum policy is instructed to reason as a transportation engineer and to search for gaps across the principal stages of a simulation study. Third, the action and critic agents are given SUMO-specific responsibilities: the action agent must execute native tools and produce simulation artifacts, while the critic checks simulator warnings, output validity, and the completeness of multi-stage analyses.

The action space deliberately consists of SUMO's native command-line and TraCI interfaces rather than a fixed catalog of wrapper tools, as in SUMO-MCP~\citep{ye2025sumo}. The action agent can write Python or shell scripts, invoke SUMO command-line programs, inspect XML and output files, and control a running simulation through TraCI. We do not require an additional Model Context Protocol (MCP) layer because direct access avoids committing the learned skills to a particular wrapper design and leaves the agent free to use newly discovered SUMO functionality. MCP remains compatible with the framework and may be advantageous when the same abstract operation must be mapped to multiple simulators or when tools are distributed across machines; it is not required for self-evolution.

The system can begin with empty procedural and semantic stores, although a small seed collection accelerates early exploration. In our implementation, the seed covers only foundational operations, such as creating an isolated intersection, generating random demand, running a simulation, and accessing vehicle state, together with basic knowledge of SUMO and TraCI APIs. It encodes neither predefined workflows nor a fixed progression of tasks. From this starting point, coverage is expanded by the same curriculum, verification, and consolidation mechanisms. Consequently, the SUMO-specific content resides primarily in the accumulated memory, while the self-evolving mechanism remains applicable to other executable domains that provide tools, observable outcomes, and reusable structure.

\section{Case Studies}
\label{sec:case_studies}

This section presents representative episodes from SimSkill's autonomous learning process. Together, they demonstrate curriculum construction, memory retrieval and task decomposition, revision through criticism, learning from both successful and failed tasks, and memory maintenance. The snapshot considered here follows approximately 80 hours of autonomous operation over five days on a MacBook Pro with an M1 Max processor, using \texttt{claude-opus-5} as the principal backbone. Starting from 5 seed skills and 10 seed knowledge pages, SimSkill accumulated 150 procedural skills and 153 semantic-memory pages. During this process, it has completed 54 memory-lint passes, and there have been 60 procedural and 201 semantic revision events, indicating that memory growth involved substantial revision and maintenance as well as artifact creation.

The accumulated procedural skills and semantic knowledge pages span the principal stages of a traffic-simulation study and range from primitive operations to composed multi-step procedures. Table~\ref{tab:skill_coverage} categorizes all 150 skills and 153 knowledge pages by primary function. Each artifact is counted once for summary purposes, although many span category boundaries and explicitly link to artifacts in other groups. The complete name-level inventory is provided in Listing~\ref{lst:complete_skill_inventory} in Appendix~\ref{app:skill_inventory}.

\begin{table*}[t]
    \centering
    \small
    \caption{Summary of the accumulated memory after approximately 80 hours of autonomous learning: 150 procedural skills and 153 semantic knowledge pages, grouped by primary function.}
    \label{tab:skill_coverage}
    \begin{tabular}{p{0.25\textwidth}rrp{0.47\textwidth}}
        \hline
        Primary function & Pages & Skills & Scope \\
        \hline
        Scenario construction, execution, and vehicle-state operations & 6 & 6 & Simulation runners, numerical configuration, output inspection, and vehicle-state querying and control. \\
        Network and infrastructure design & 11 & 12 & Synthetic and imported networks, geometric facilities, subnetworks, lane permissions, and format fidelity. \\
        Demand, routing, and assignment & 17 & 17 & Demand generation, OD conversion, assignment, route choice, equilibrium, and count- or trajectory-based reconstruction. \\
        Signals and intersection control & 31 & 30 & Signal timing, adaptive and predictive control, progression, priority, intersection treatments, and roundabout control. \\
        Freeway, corridor, and network operations & 31 & 31 & Interchanges, work zones, ramp control, managed facilities, incidents, pricing, resilience, and network management. \\
        Transit, multimodal, fleet, and parking systems & 22 & 20 & Transit planning and operations, rail, walking, cycling, parking, freight, taxis, and electrification. \\
        Calibration, estimation, and experimental design & 22 & 21 & Behavioral and demand calibration, state and capacity estimation, sensitivity, stochastic replication, forecasting, and analytical validation. \\
        Impact analysis, validation, and visualization & 13 & 13 & Safety, environment, equity, economics, reliability, artifact diagnosis, and result communication. \\
        \hline
        Total & 153 & 150 & Complete memory snapshot. \\
        \hline
    \end{tabular}
\end{table*}

\subsection{Curriculum Construction, Memory Retrieval, and Task Decomposition}
\label{sec:case_curriculum}

The curriculum agent searches for missing capabilities and decision variables rather than merely for absent topic names. For example, the memory already contained methods for estimating an origin--destination (OD) matrix from traffic counts, but every method treated count locations as fixed. The agent therefore proposed count-station placement as a new decision problem, comparing random, volume-greedy, rule-based, observability-based, and D-optimal designs across sensor budgets. In another iteration, it found that the signal-control collection contained offline and reactive controllers but no method that optimized future switching decisions against predicted arrivals. That gap led to the predictive-control episode in Section~\ref{sec:case_predictive}.

In the current implementation, SimSkill's metadata-first retrieval policy is executed with Claude Code's built-in file search rather than a purpose-built retrieval component. For the infrastructure-planning task PP-T4-3-V2 from Benchmark V2 (Section~\ref{sec:experiments}), the retriever selected eight procedural skills and two semantic pages covering grid construction, OD demand, transit, accessibility and equity, benefit--cost analysis, network design, and simulation execution. The action agent then decomposed the task into network construction, demand generation, experimental design, simulation, stress testing, analysis, and appraisal. No retrieved artifact solved the complete task; the retrieved set supplied complementary components from which the agent constructed a task-specific workflow. An abridged trace of the retrieved memories and decomposition is given in Listing~\ref{lst:pp_retrieval_trace} in Appendix~\ref{app:learning_traces}.

In the observed 150-skill, 153-page memory collection, Claude Code's built-in file search did not become a bottleneck. This is an empirical observation at the present scale, not a claim of indefinite scalability: a library containing tens of thousands of artifacts may require hierarchical indexing, learned retrieval, or domain-specific routing. SimSkill separates retrieval policy from memory representation, allowing the retriever to be replaced without changing the stored skills, knowledge pages, or learning loop.

\subsection{Learning from an Unsuccessful Task}
\label{sec:case_failure}

A station-based shared-micromobility learning episode shows why task failure need not imply learning failure. The curriculum agent detected that memory lacked any account of bike-sharing simulation, and further inferred that SUMO lacked the required primitive. The action agent attempted three times; each attempt was rejected by the critic. Table~\ref{tab:micromobility_retractions} summarizes the progression.

\begin{table*}[t]
    \centering
    \small
    \caption{Successive revisions during the shared-micromobility episode.}
    \label{tab:micromobility_retractions}
    \begin{tabular}{p{0.07\textwidth}p{0.34\textwidth}p{0.50\textwidth}}
        \hline
        Attempt & Action agent's conclusion & Critic agent's verdict \\
        \hline
        1 & Bicycle trips appear to create bicycles automatically, so SUMO cannot enforce a finite shared-fleet inventory. & Defining a fixed set of named bicycle vehicles and using person-triggered stops restricts service to that fleet; additional travelers are denied once all bicycles are in use. \\
        2 & A traveler must request a specific bicycle in advance and therefore cannot use whichever bicycle is available. & The \texttt{lines="ANY"} mechanism assigns any eligible bicycle at runtime. The earlier failure was caused by invalid candidate routes whose errors had been suppressed. \\
        3 & A bicycle can either wait at a station as available inventory or carry a traveler, but cannot transition between these roles. & With the bicycle's passenger capacity set correctly, a time-bounded person-triggered stop allows the same bicycle to unload one traveler, remain available at the station, board a later traveler, and then depart with that traveler. \\
        \hline
    \end{tabular}
\end{table*}

Because the motivating claim did not survive verification, SimSkill created no procedural skill. It instead created the semantic page \texttt{station-based-shared-micromobility-in-sumo}, which records the measured boundary between native functionality and the residual need for external control logic. It also corrected two existing knowledge pages concerning stop-output fields and offline observation of parking-area occupancy. The reusable outcome was therefore a better account of what the environment supports.

The episode also shows the agent revising its own curriculum policy. SimSkill distilled the principle that ``memory is silent about $X$'' implies only that $X$ is not covered by memory, not that the environment cannot provide $X$. Later tasks in the same Claude Code session that were premised on a missing simulator capability were then required to challenge that premise through documentation and source inspection. The agent also relaxed its novelty bias: ``characterization, validation, comparison, and integration can yield valuable learning tasks without first asserting a simulator deficiency.'' Its next demand-reconstruction task accordingly verified that SUMO provides \texttt{tracemapper.py} and \texttt{route2OD.py} before identifying the genuine gap---the memory covered reconstruction from fixed-point counts but not from sparse GPS trajectories. Appendix~\ref{app:learning_traces} preserves the corresponding curriculum trace in Listing~\ref{lst:curriculum_revision_trace}.

These newly discovered principles are not made persistent; they survive only within the Claude Code session in which they arise. Adding a mechanism for updating the system skills themselves would preserve them, allowing a self-evolving system to revise not only its memory but also the rules and principles by which it operates. This is a form of meta-learning---learning how to learn---which we leave to future work.

\subsection{Acquisition of a Predictive-Control Skill}
\label{sec:case_predictive}

The predictive rolling-horizon signal-control episode, recorded as the acquisition of the 147th learned skill, illustrates the complete learning loop. The curriculum agent identified anticipatory control as a gap in a library of offline and reactive controllers. Retrieval then supplied adjacent capabilities---closed-loop control, rolling-horizon forecasting, detector design, baseline signal timing, and stochastic evaluation---which the action agent composed into a new solution.

The first action-agent attempt produced executable artifacts and a complete report, but the critic rejected it because key claims were insufficiently verified and the predictive controller had been evaluated only with an arbitrary 8-second minimum-green setting. This setting caused the controller to switch at the earliest permitted time in most decisions, so the measured performance reflected the chosen constraint rather than the predictive method alone. The critic's findings were passed to a fresh action-agent iteration, whereas local timeout recovery and script-debugging details remained isolated within the first attempt. After revising the experimental design and independently validating the critical claims, the second attempt was accepted. The episode thus demonstrates the division between the inner execution loop, which repairs operational errors, and the outer action--critic loop, which addresses conceptual and methodological errors.

Although the study did not establish an advantage for predictive control, it yielded reusable procedures, validation rules, and knowledge about the method's operating boundary. Ingestion therefore created a new skill and semantic page while revising related artifacts whose assumptions had been tested more rigorously, as summarized in Table~\ref{tab:predictive_memory_changes}. After ingestion, an incremental memory-lint pass verified the changed artifacts, bundled scripts, and cross-references against the full collection. The more detailed trace is provided in Listing~\ref{lst:predictive_learning_trace} in Appendix~\ref{app:learning_traces}.

\begin{table*}[t]
    \centering
    \small
    \caption{Memory changes produced by the predictive-control learning episode.}
    \label{tab:predictive_memory_changes}
    \begin{tabular}{p{0.15\textwidth}p{0.25\textwidth}p{0.52\textwidth}}
        \hline
        Memory operation & Artifact & Content added or corrected \\
        \hline
        New skill & \texttt{implement-predictive-}\newline
        \texttt{rolling-horizon-signal-}\newline
        \texttt{control} & A reusable procedure for implementing and validating predictive rolling-horizon signal control, supported by three executable scripts. \\
        Skill update & \texttt{build-rolling-horizon-}\newline
        \texttt{traffic-forecast-with-}\newline
        \texttt{state-warm-start} & Clarified when restoring a saved simulation state can duplicate demand and how the required safeguards differ between TraCI and command-line execution. \\
        Skill update & \texttt{optimize-under-simulation-}\newline
        \texttt{noise-with-a-fixed-budget} & Added a requirement to test key controller settings across plausible values before comparing controllers, so that the result does not depend on a single arbitrary configuration. \\
        New knowledge page & \texttt{value-of-anticipation-in-}\newline
        \texttt{predictive-signal-control} & Summarized when advance information about vehicle arrivals improves signal control and when prediction error, congestion, or computational cost removes that benefit. \\
        Knowledge page update & \texttt{state-serialization-and-}\newline
        \texttt{rolling-horizon-traffic-}\newline
        \texttt{forecasting} & Documented when restored vehicle-flow schedules can duplicate future departures, and showed that TraCI may reset the reported elapsed time of the active signal phase after state loading, requiring controllers to track this time externally. \\
        \hline
    \end{tabular}
\end{table*}

\subsection{Compositional Skill Acquisition and Consistent Memory Repair}
\label{sec:case_composition}

The GPS map-matching learning episode illustrates how SimSkill acquires a capability by composing existing ones. The action agent combined procedures for network import, demand reconstruction, stochastic evaluation, and output analysis to produce \texttt{map-match-gps-traces-to-reconstruct-demand}, together with four reusable scripts. The new skill was therefore built from accumulated procedural memory rather than developed as an isolated solution.

Composition also tested an older dependency in a new setting. A reused script from \texttt{load-osm-network} failed because of incorrect command-line argument handling. The action agent diagnosed and repaired the defect, after which ingestion updated the dependency together with the new skill and related semantic knowledge. Table~\ref{tab:map_matching_memory_changes} summarizes this coordinated transaction.

\begin{table*}[t]
    \centering
    \small
    \caption{Memory changes produced by the GPS map-matching learning episode.}
    \label{tab:map_matching_memory_changes}
    \begin{tabular}{p{0.15\textwidth}p{0.22\textwidth}p{0.53\textwidth}}
        \hline
        Memory operation & Artifact & Content added or corrected \\
        \hline
        New skill & \texttt{map-match-gps-traces-}\newline
        \texttt{to-reconstruct-demand} & A reusable procedure for converting sparse GPS traces into routes and OD demand, supported by four scripts for probe generation, map matching, evaluation, and OD aggregation. \\
        Skill repair & \texttt{load-osm-network} & Fixed a wrapper-script error that prevented OSM downloads for bounding boxes containing negative longitudes, and corrected the corresponding instructions. \\
        New knowledge page & \texttt{gps-map-matching-and-}\newline
        \texttt{probe-demand-}\newline
        \texttt{reconstruction} & Summarized how sampling interval, location error, probe penetration, missing observations, and matcher settings affect reconstruction accuracy. \\
        Knowledge page update & \texttt{geh-statistic} & Added evidence that a reconstruction can fit aggregate link counts while still recovering routes or OD demand poorly, motivating the use of complementary accuracy measures. \\
        \hline
    \end{tabular}
\end{table*}

This case shows that skill reuse serves both composition and continued validation. New capabilities can build on previously learned scripts, while failures encountered during reuse provide evidence for correcting earlier memory. Because the new skill, repaired dependency, and semantic updates were committed together, subsequent agents receive a consistent set of artifacts. A more detailed trace is provided in Listing~\ref{lst:map_matching_learning_trace} in Appendix~\ref{app:learning_traces}.

\subsection{Memory Maintenance as Library-Level Learning}
\label{sec:case_lint}

As memory grows, episode-level ingestion alone cannot guarantee the health of the collection. New artifacts may overlap or conflict with existing ones, references and interfaces may become outdated, and errors in bundled scripts may silently propagate through later skill composition. \texttt{memory-lint} therefore performs collection-level quality control: it checks recent changes against existing memory, repairs inconsistencies, maintains cross-references, and identifies redundant or obsolete content. Table~\ref{tab:lint_cases} presents representative examples.

\begin{table*}[t]
    \centering
    \small
    \caption{Representative memory-lint operations for maintaining memory health.}
    \label{tab:lint_cases}
    \begin{tabular}{p{0.20\textwidth}p{0.40\textwidth}p{0.31\textwidth}}
        \hline
        Memory-health issue & Observed evidence & Maintenance action \\
        \hline
        Broken, drifted, and potentially duplicate artifacts & \texttt{fcd-postprocessing} linked to a nonexistent page; \texttt{actuated-signal-control} used the obsolete \texttt{--sumo-seed} argument; two intersection-construction skills overlapped. & Repaired the page and argument references. The possible merge was recorded but deferred because equivalence had not been established. \\
        Executable metric bug & Two scripts summed a cumulative teleport counter over all time steps, silently overcounting whenever teleports occurred. & Changed both scripts to read the final cumulative value and corrected the associated output-format knowledge page. \\
        Broken skill reference & The \texttt{implement-glosa-speed-}\newline \texttt{advisory-controller} skill referred to \texttt{change-vehicle-state}, which is a knowledge page; the intended procedural skill is \texttt{set-vehicle-state}. & Replaced the incorrect reference with \texttt{set-vehicle-state} and verified the links between the skill and the related knowledge page. \\
        Missing reciprocal skill link & The Braess-paradox skill linked to the departure-time-equilibrium skill, but the departure-time skill did not link back. Both concern equilibrium effects arising from travelers' choices: route choice in one case and departure-time choice in the other. & Added the missing backlink from the departure-time skill to the Braess-paradox skill. \\
        Internal inconsistency & The saturation-flow skill initially recommended one estimator in all cases, but a later passage explained that this estimator becomes unreliable with deterministic driver behavior. & Rewrote the initial recommendation so that the choice of estimator depends on whether driver behavior is stochastic or deterministic. \\
        Incorrect reference & A procedural skill cited the \texttt{sumo-output-files} knowledge page using the notation for a skill name, rather than the semantic-memory link \texttt{[[sumo-output-files]]}. & Corrected only the citation syntax so that the reference explicitly points to the semantic-memory page. \\
        Related skills flagged as possible duplicates & A similarity check paired predictive rolling-horizon control with max-pressure control, and count-station design with OD estimation. However, the paired skills solve different problems: predictive versus reactive signal control, and sensor placement versus demand estimation. & Kept the skills separate after confirming their distinct purposes and verifying their scripts and links. \\
        \hline
    \end{tabular}
\end{table*}

Memory linting is essential because the usefulness of accumulated experience depends on its continued trustworthiness. It keeps memory correct by repairing content and executable defects, up to date by replacing drifted names and interfaces, connected by maintaining links and indexes, and relevant by controlling redundancy. When the evidence for a change is insufficient, the issue is recorded and deferred rather than resolved speculatively. Every check and modification is retained in log file (\texttt{log.md} in the public repository), making the maintenance process auditable. Memory maintenance is therefore part of learning itself: it preserves a reliable foundation on which later retrieval and skill composition can operate.

\subsection{Flexibility, Reliability Limits, and Safeguards of Language-Specified Orchestration}
\label{sec:case_language_control}

In SimSkill, the high-level learning workflow is specified through natural-language system skills rather than a hard-coded control program. This representation makes the workflow transparent, editable, and easy to extend as the system evolves. The same flexibility introduces a reliability limitation: natural-language instructions are not enforced as deterministically as program control flow. For example, in pilot runs with a weaker orchestrator (\texttt{claude-haiku}), a polished success report from the action agent was occasionally read as completion of the entire workflow, and the orchestrator stopped before critic evaluation, episodic recording, or memory ingestion.

Language-based orchestration can therefore be combined with explicit safeguards at points where an omitted step would compromise inference or learning. An external completeness check, for example, can verify that all mandatory artifacts have been produced and, if they have not, resume the session with a prompt to complete the remaining steps. Such safeguards are most valuable with weaker backbone models and wherever a critical workflow requirement needs a deterministic check.

\section{Experiments}
\label{sec:experiments}

We evaluate whether the explicit experience accumulated by SimSkill improves its ability to solve held-out traffic-simulation tasks. The experiment compares the complete system with both memory-ablated variants and a plain Claude Code baseline, while varying the backbone LLM and task difficulty. It addresses the following research questions:

\begin{enumerate}
    \item[\textbf{RQ1}] \emph{Task completion:} Does the complete SimSkill system solve more tasks than the same inference framework without accumulated memory and than vanilla Claude Code?
    \item[\textbf{RQ2}] \emph{Backbone dependence:} Are the effects consistent across backbone LLMs, or do they depend on how well a model interprets and executes the framework and on its compatibility with Claude Code?
    \item[\textbf{RQ3}] \emph{Memory contribution:} How much do procedural memory and semantic memory contribute individually and jointly, relative to the inference framework alone?
    \item[\textbf{RQ4}] \emph{Resource efficiency:} How does memory change the trade-off between verified task success, monetary cost, and wall-clock time?
    \item[\textbf{RQ5}] \emph{Generalization:} Do the benefits persist on held-out compositional and novel tasks rather than only on tasks closely resembling stored experience?
\end{enumerate}

\subsection{Experimental Setup}
\label{sec:experiment_setup}

\paragraph{Benchmarks.}
We constructed two frozen benchmarks, each comprising 40 standalone tasks. Benchmark V1 spans ten capability areas covering the full simulation workflow, including network generation, demand generation, signal control and optimization, TraCI-based closed-loop control, and output analysis and visualization. Its 40 tasks are evenly distributed across four difficulty tiers, with 10 tasks per tier; higher tiers require progressively more complex reasoning and stronger generalization.

Benchmark V2 concentrates on substantially more difficult tasks. It contains 40 tasks across network design, demand inference, signal control, freeway operations, multimodal transit, safety and human factors, environment and energy, freight and curb operations, planning and policy equity, and simulation methodology. It is divided into 20 Tier~3 and 20 Tier~4 tasks, where Tier~4 demands more generalization than Tier~3. Both V2 tiers are substantially harder than their V1 counterparts.

Five V1 tasks and ten V2 tasks include accompanying input files, such as detector counts, origin--destination matrices, trajectories, and General Transit Feed Specification (GTFS) data. Each prompt provides all other information needed for execution and verification. Listing~\ref{lst:example_benchmark_tasks} presents four representative V1 tasks---one per tier---and two representative V2 tasks.

\lstdefinestyle{benchmarktask}{
    basicstyle=\ttfamily\scriptsize,
    keywordstyle=\bfseries\color{blue!55!black},
    morekeywords={Benchmark,V1,V2,Tier,Input},
    breaklines=true,
    breakatwhitespace=true,
    breakindent=1em,
    columns=fullflexible,
    keepspaces=true,
    showstringspaces=false,
    frame=single,
    framerule=0.3pt,
    rulecolor=\color{black!25},
    backgroundcolor=\color{black!2},
    framesep=3pt,
    xleftmargin=0.2em,
    xrightmargin=0.2em,
    aboveskip=0.5em,
    belowskip=0.5em,
    captionpos=t
}

\begin{lstlisting}[style=benchmarktask,
                   caption={Example tasks from Benchmarks V1 and V2.},
                   label={lst:example_benchmark_tasks}]
[Benchmark V1 | NG-T1 | network-generation | Tier 1]
Build a single 4-legged intersection in SUMO with traffic-light control on all approaches, each approach 2 lanes wide with a 300 m link length.

[Benchmark V1 | SC-T2 | signal-control-optimization | Tier 2]
Coordinate three consecutive signals along a one-way arterial (signals spaced 400 m apart, posted speed limit 50 km/h) into a green wave using tlsCoordinator, at a progression speed 15% below the posted speed limit (42.5 km/h). Use the same 60 s cycle at each signal, comprising 27 s arterial green, 3 s yellow, and 30 s red.

[Benchmark V1 | MT-T3 | multimodal-transit | Tier 3]
Import the GTFS feed in test/benchmark_tasks_files/MT-T3/gtfs/ for a small single-route transit network (6 stops, hourly service). Place a signalized cross street immediately upstream of stop S4. Use a 60 s fixed-time signal with 25 s arterial green, 25 s cross-street green, 3 s yellow, and 2 s all-red. Add background demand of 600 veh/h on the arterial and 300 veh/h on the cross street. Compare this baseline with transit signal priority at S4. Quantify the resulting schedule-adherence improvement relative to no priority.
Input: test/benchmark_tasks_files/MT-T3/gtfs/

[Benchmark V1 | CA-T4 | calibration | Tier 4]
Reproduce the classical 4-node Braess's-paradox network in SUMO: a start node S and end node E connected via two parallel routes through intermediate nodes A and B, where edges S->A and B->E have travel time that increases with flow (capacity-sensitive -- 2 km at 50 km/h free-flow, congesting under load) while edges S->B and A->E have a fixed travel time independent of flow (a 3-minute fixed-time edge each), plus a new low-cost bridging edge A->B. Empirically verify, under a fixed demand of 4000 veh/h from S to E, whether adding the A->B bridge makes average travel time worse under dynamic user equilibrium, rather than assuming the classical result transfers unchanged.

[Benchmark V2 | NG-T3-1-V2 | network-design | Tier 3]
Build a skewed four-leg signalized intersection from explicit node, edge, and connection files rather than a stock generator. Each arm is 400 m long at 50 km/h; outward bearings counter-clockwise from east are east (E)=0, north (N)=78, west (W)=183, and south (S)=266 degrees. Upstream inbound/outbound lane counts are N=1/2, E=2/3, S=3/2, and W=2/1. Flare the north inbound to two stop-line lanes with a 90 m left bay; the east inbound to four lanes with a 120 m left bay and 80 m right bay; and the south inbound to four lanes with a 70 m right bay. Stop-line assignments are N=[left, through+right], E=[left, through, through, right], S=[left, through, through, right], and W=[left+through, through+right]. Serve all 12 non-U-turn movements, give each left a protected interval, and use 3 s yellow plus 2 s all-red. Supply the lane-level movement table, signal-link map, and an automated test that compiles without warnings, routes one probe through every legal movement, rejects every U-turn, detects unintended lane permissions, and fails on any teleport or missing signal link. Put all legal probes in one simulation and all deliberately illegal probes in a second. Use one fixed random seed for both executions.

[Benchmark V2 | PP-T4-3-V2 | planning-policy-equity | Tier 4]
Estimate long-run induced demand and relocation after adding a 2 km, two-lane bypass to a four-zone town, rather than holding land use and trip totals fixed. Represent the town as a 3x3 grid with 500 m blocks, one 50 km/h lane per direction, signalized interior junctions, zones 1-4 at the northwest (NW), northeast (NE), southwest (SW), and southeast (SE) fringe, and the central business district (CBD) at the center. Connect both southern zone centroids to the south-center node; the proposed 70 km/h bypass links that node to the east-center node and has no intermediate access. Base households by zone are [1200,900,700,600], jobs [400,1100,800,500], peak car trips 4200, and observed free-flow zone-to-center times [22,18,27,31] minutes. Add nonnegative centroid-connector access times so each simulated grid path plus connector time reproduces its listed zone-to-center value; hold those access times fixed in every year and alternative. Construct the base directed origin-destination (OD) matrix with a doubly constrained gravity model whose productions are proportional to households, attractions are proportional to jobs, and impedance is exp(-0.08*t_ij): for different zones set t_ij to the sum of their two listed center times, and for an intrazonal cell use half that zone's listed time. Include intrazonal trips and scale all cells to exactly 4200 trips. The bypass costs $18 million and initially saves eight minutes for zones 3-4. Simulate 10 annual iterations: trip frequency elasticity to generalized time is -0.35; 4% of households and 6% of jobs may relocate each year using a logit response with time coefficient -0.08/min and fixed zone capacities of [1500,1200,1200,900] households and [700,1400,1200,900] jobs. Each year rebalance trips, use SUMO's native rerouting device within one annual execution, update accessibility from completed trips, and apply one damped relocation update with damping 0.5. Run all 10 years, but also record the first year after which both travel times and locations change below 1% for two successive annual updates. Compare no-build and bypass with one fixed traffic seed per annual state. Report generated vehicle-kilometers traveled (VKT), congestion rebound, accessibility, relocations, carbon dioxide (CO2), and whether the initial benefit persists; test elasticities -0.15 and -0.55 and distinguish modeled behavioral assumptions from simulated traffic outcomes. Implement gravity balancing, elasticity, relocation, damping, capacity enforcement, and annual state transfer in task-owned code. Evaluate elasticities -0.15 and -0.55 only as first-order offline trip-total sensitivities using the cached annual generalized times, and label them unsimulated.
\end{lstlisting}

\paragraph{Baseline and ablations.}
Table~\ref{tab:experiment_conditions} defines the five conditions. The headline comparison uses the complete system (\texttt{full-ver}) and plain Claude Code (\texttt{vanilla-cc}) on both benchmarks with DeepSeek-V4-Pro, GLM-5.2, and Qwen3.7-Max. The five-way ablation is conducted on Benchmark V1 with DeepSeek-V4-Pro and Qwen3.7-Max. In every SimSkill condition, the orchestrator and its sub-agents use the same backbone model.

\begin{table*}[t]
    \centering
    \small
    \setlength{\tabcolsep}{4pt}
    \caption{Experimental conditions. The four SimSkill conditions form a $2\times2$ design over procedural and semantic memory; the vanilla condition additionally removes the SimSkill inference framework.}
    \label{tab:experiment_conditions}
    \begin{tabular}{p{0.16\textwidth}p{0.10\textwidth}p{0.12\textwidth}p{0.11\textwidth}p{0.40\textwidth}}
        \hline
        Condition & Infer framework & Procedural memory & Semantic memory & Interpretation \\
        \hline
        \texttt{full-ver} & Yes & Yes & Yes & Complete SimSkill system. \\
        \texttt{proc-mem-ver} & Yes & Yes & No & Isolates procedural skills while retaining the action--critic architecture. \\
        \texttt{sem-mem-ver} & Yes & No & Yes & Isolates declarative knowledge while retaining the action--critic architecture. \\
        \texttt{infer-frame-only} & Yes & No & No & Retains retrieval control, sub-agents, criticism, and retry logic, but removes accumulated memory content. \\
        \texttt{vanilla-cc} & No & No & No & Removes \texttt{CLAUDE.md} and the SimSkill system skills and sub-agents; Claude Code receives the task only. \\
        \hline
    \end{tabular}
\end{table*}

\paragraph{Execution protocol.}
Each condition--model batch was instantiated in a disposable Git worktree. The appropriate memory directories and framework files were retained, emptied, or removed only inside that worktree. Every task was then run in a fresh, cold-context model session. SimSkill was invoked in test mode: it retrieved memory, executed the action--critic loop, and saved an episodic record, but did not ingest the benchmark experience into procedural or semantic memory. The vanilla condition was likewise required to save its artifacts in the common episodic format so that all outputs could be evaluated identically. At the end of a batch, the worktree was discarded; no evaluation artifact was merged into the memory under test.

\paragraph{Independent evaluation.}
After a task session ended, Claude Opus~5 was launched as an independent judge in a separate Claude Code session and worktree with no prior context. It received the original task text and the saved episodic memory, reran simulations, and checked reproducibility, scope, and numerical claims. For the V2 DeepSeek-V4-Pro and Qwen3.7-Max runs, we additionally used GLM-5.2 as an independent pointwise judge, which decomposed each task into weighted requirements and assigned a completion score in $[0,1]$, where zero denotes no verifiable task-specific result and one a complete and correct solution. This continuous score captures substantial partial progress that a binary verdict necessarily discards.

\paragraph{Cost and time metrics.}
We report the task-solving session's dollar cost and total wall-clock time. Input, output, cache-creation, and cache-read tokens were repriced with a fixed provider-specific table dated 17 August 2026. The wall-clock metric includes both model/tool orchestration and simulator execution and therefore represents user-observed elapsed time. For condition $c$, let $\mathcal{E}_c$ be the runs with both a conclusive independent verdict and an observed value of the budget metric. The plotted success-at-budget curve is
\begin{equation}
    A_c(b)=\frac{1}{|\mathcal{E}_c|}\sum_{i\in\mathcal{E}_c}
    \mathbf{1}\!\left[V_{ci}=1 \;\land\; B_{ci}\leq b\right],
    \label{eq:success_at_budget}
\end{equation}
where $V_{ci}$ is the independent verdict and $B_{ci}$ is either observed cost or wall-clock time. A verified failure remains in the denominator and never increments the curve. Median cost and time are also reported, but they are secondary: a condition that fails quickly can have a deceptively small median.

\subsection{Main Results}
\label{sec:experiment_results}

\begin{figure}[H]
    \centering
    \includegraphics[width=0.94\textwidth]{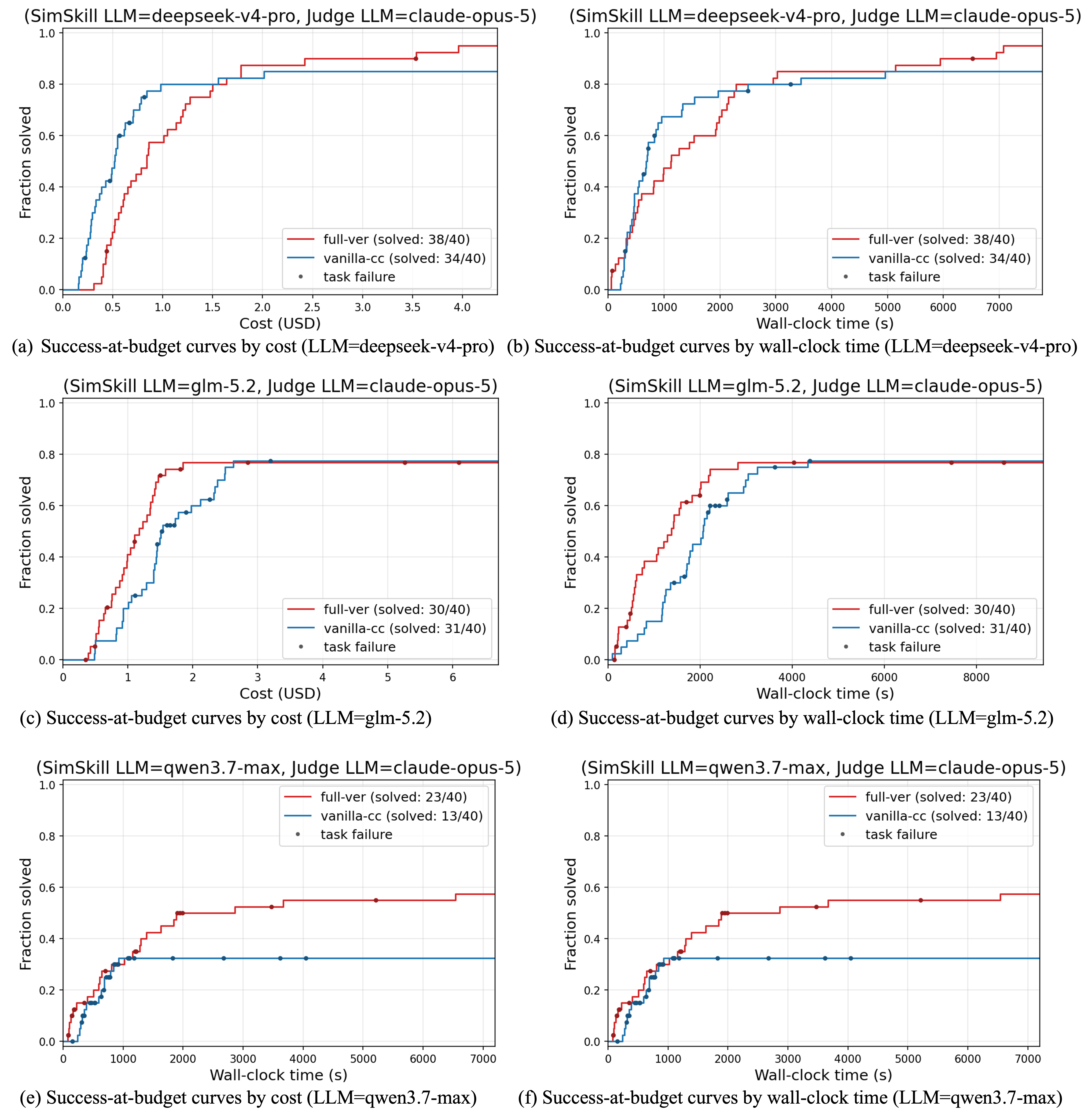}
    \caption{Verified task success on Benchmark V1 as a function of observed monetary or wall-clock budget. Each panel uses the budget shown on its horizontal axis. Red curves show complete SimSkill and blue curves show vanilla Claude Code; endpoint labels give verified success out of all 40 tasks, and markers locate verified failures at their consumed resource levels.}
    \label{fig:performance_benchmark_v1}
\end{figure}

\begin{figure}[H]
    \centering
    \includegraphics[width=0.94\textwidth]{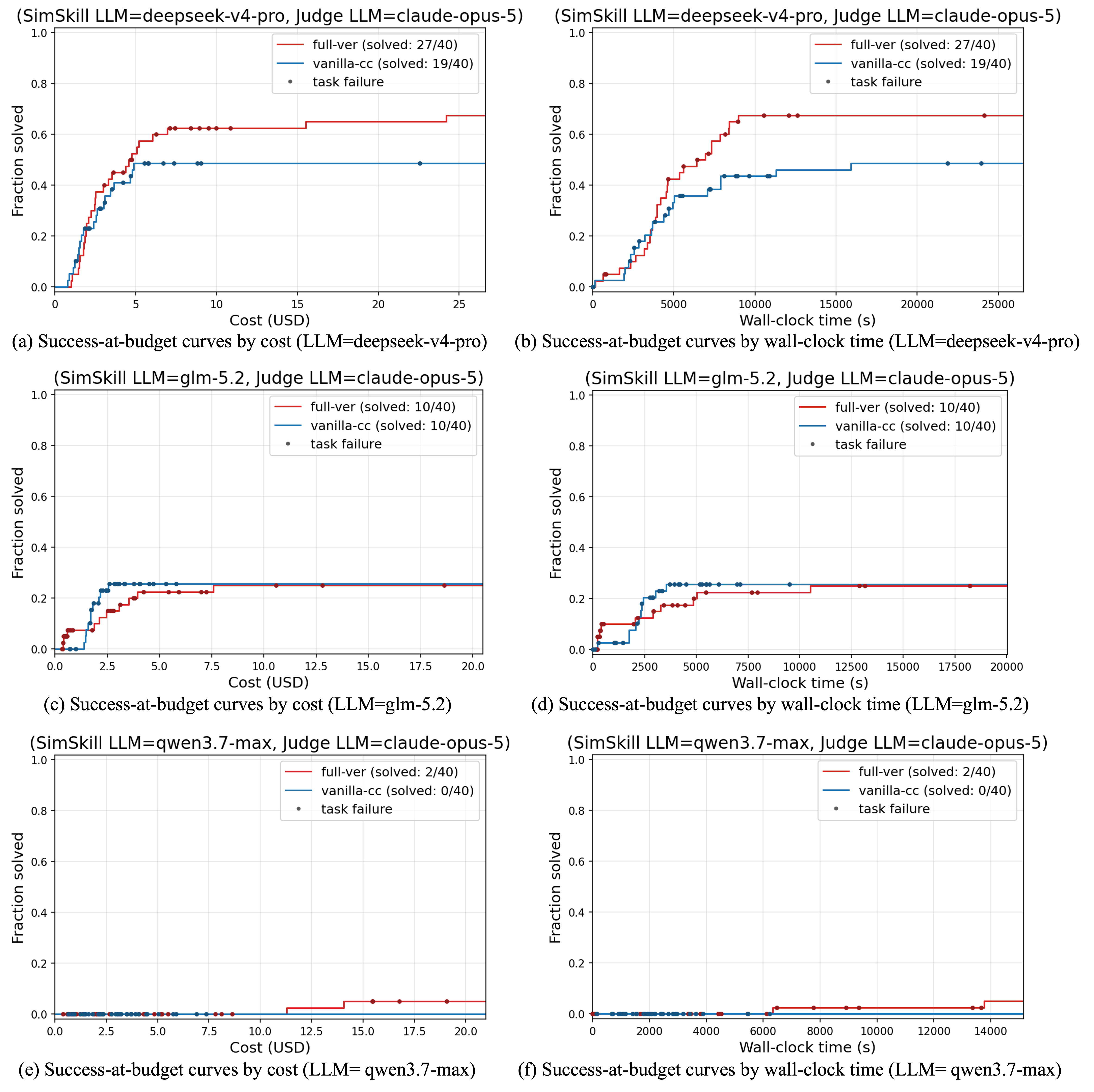}
    \caption{Verified task success on the hard Benchmark V2 as a function of observed monetary or wall-clock budget. Plot semantics are the same as in Figure~\ref{fig:performance_benchmark_v1}.}
    \label{fig:performance_benchmark_v2}
\end{figure}

Table~\ref{tab:headline_results} compares the task-solving performance of \texttt{full-ver} and \texttt{vanilla-cc} on Benchmarks V1 and V2 across the three backbones, together with median per-run cost (USD) and wall-clock time (s). Figures~\ref{fig:performance_benchmark_v1} and~\ref{fig:performance_benchmark_v2} show the corresponding empirical success-at-budget curves. The curves expose both eventual coverage and the budget required to reach it; markers denote runs that consumed resources but failed independent verification.

\begin{table*}[t]
    \centering
    \small
    \setlength{\tabcolsep}{4pt}
    \caption{Complete SimSkill versus vanilla Claude Code. Verified successes out of 40 tasks are reported, with percentages in parentheses. Cost and time entries are per-run medians in the form full (F)/vanilla (V).}
    \label{tab:headline_results}
    \begin{tabular}{llccrrr}
        \hline
        Benchmark & Backbone & \texttt{full-ver} & \texttt{vanilla-cc} & $\Delta$ (pp) & Cost F/V (USD) & Time F/V (s) \\
        \hline
        V1 & DeepSeek-V4-Pro & 38 (95.0\%) & 34 (85.0\%) & $+10.0$ & 0.78/0.49 & 1056/650 \\
        V1 & GLM-5.2          & 30 (75.0\%) & 31 (77.5\%) & $-2.5$  & 1.02/1.47 & 918/1918 \\
        V1 & Qwen3.7-Max      & 23 (57.5\%) & 13 (32.5\%) & $+25.0$ & 1.84/1.03 & 851/683 \\
        V2 & DeepSeek-V4-Pro & 27 (67.5\%) & 19 (47.5\%) & $+20.0$ & 3.93/2.92 & 4623/4796 \\
        V2 & GLM-5.2          & 10 (25.0\%) & 10 (25.0\%) & $0.0$   & 2.29/2.44 & 2008/2899 \\
        V2 & Qwen3.7-Max      &  2 (5.0\%)  &  0 (0.0\%)  & $+5.0$  & 2.16/2.35 & 916/1969 \\
        \hline
    \end{tabular}
\end{table*}

With DeepSeek-V4-Pro, SimSkill improves verified success from 34 to 38 tasks on V1 and from 19 to 27 tasks on V2, corresponding to absolute gains of 10 and 20 percentage points. Qwen3.7-Max exhibits the largest V1 gain: 23 tasks are verified under SimSkill compared with 13 under vanilla Claude Code. On V2, vanilla Qwen3.7-Max solves no task completely, whereas SimSkill solves two. SimSkill therefore enables some long-horizon completion even when the baseline does not, although the absolute V2 result of 2/40 also shows that scaffolding and memory cannot compensate fully for limitations of the backbone.

\paragraph{Backbone dependence.} This improvement is not universal across backbones. GLM-5.2 shows no gain: SimSkill even trails vanilla by one task on V1 and ties it on V2. SimSkill's benefit therefore depends on the backbone: how reliably it follows long-horizon control instructions, uses tools, delegates to sub-agents, and acts on retrieved material, and how compatible it is with Claude Code.

\paragraph{Generalization to difficult tasks.}
The gains are not confined to tasks solvable by a direct skill call. On V1, Qwen3.7-Max improves in every tier, including a gain of four tasks among the 10 Tier~4 tasks. On V2, DeepSeek-V4-Pro gains two tasks in Tier~3 (11/20 versus 9/20) and six in Tier~4 (16/20 versus 10/20). Both of Qwen3.7-Max's two V2 successes are Tier~4 tasks. These results provide evidence that the accumulated library can support new compositions and tasks that must be solved from first principles, rather than only near-duplicates of past experience.

Appendices~\ref{app:oa_t3}--\ref{app:mt_t4_4_v2} make this aggregate result concrete through three SimSkill task execution with DeepSeek-V4-Pro backbone. OA-T3 is a congestion visualization and safety analysis task from Benchmark V1; DG-T4-3-V2 and MT-T4-4-V2 are Benchmark V2 tasks of time-dependent OD inference and shared e-scooter operations. Together they illustrate how accumulated memory participates in inference.

\paragraph{Accuracy--resource trade-off.}
The addition of memory does not produce a uniform reduction in inference cost. On V1, vanilla Claude Code often solves the easiest tasks at smaller budgets because it avoids retrieval and multi-agent orchestration. DeepSeek-V4-Pro and Qwen3.7-Max reach higher endpoints under SimSkill, but their median costs rise from \$0.49 to \$0.78 and from \$1.03 to \$1.84, respectively. Their median times also increase. GLM-5.2 is the counterexample: despite an essentially unchanged success rate, SimSkill reduces its V1 median cost by approximately 31\% and median time by approximately 52\%, indicating that SimSkill can reduce exploration even when it does not expand the set of solvable tasks.

On V2, DeepSeek-V4-Pro pays a higher median cost under SimSkill (\$3.93 versus \$2.92) while achieving eight additional successes and a slightly lower median elapsed time. GLM-5.2 has lower SimSkill medians but no accuracy gain. The same total-cost pattern holds for DeepSeek-V4-Pro (\$203 versus \$157) and GLM-5.2 (\$128 versus \$107) on V2. Thus the completed experiment supports an accuracy--resource trade-off, not a general claim that memory makes inference cheaper. The success-at-budget curves are therefore more informative: they show how much verified task coverage is achieved at each expenditure level, whereas median cost alone may appear favorable when many runs terminate early without completing the task.

\begin{figure}[H]
    \centering
    \includegraphics[width=0.80\textwidth]{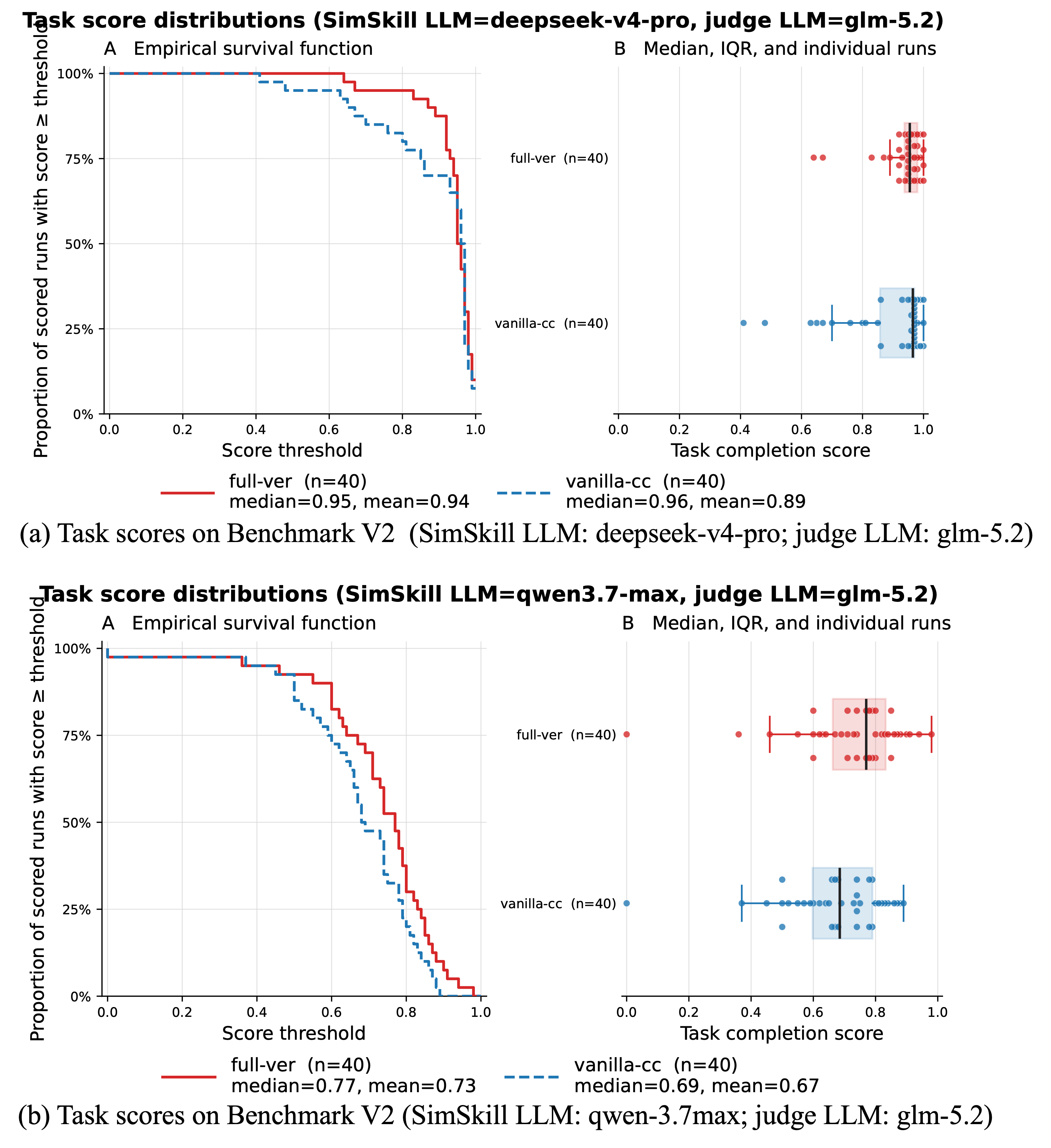}
    \caption{Scores on Benchmark V2 assigned independently by GLM-5.2. The left panels show the empirical proportion of runs whose score meets or exceeds each threshold; the right panels show individual runs together with the median and interquartile range. DeepSeek-V4-Pro is shown above and Qwen3.7-Max below.}
    \label{fig:score_distributions}
\end{figure}

\paragraph{Continuous quality scores.}
Figure~\ref{fig:score_distributions} provides a second evaluation of the Benchmark V2 outputs. Under the GLM-5.2 pointwise judge, DeepSeek-V4-Pro with SimSkill has a mean score of 0.940, compared with 0.891 for vanilla Claude Code. Its median is slightly lower (0.955 versus 0.965), showing that the improvement is concentrated in reducing the lower tail rather than uniformly shifting every task. For Qwen3.7-Max, both the mean (0.728 versus 0.674) and median (0.770 versus 0.685) favor SimSkill. The continuous judge therefore agrees with the direction of the Claude Opus~5 binary results for the two evaluated backbones while revealing substantial partial completion among most tasks that did not pass the binary threshold.

\subsection{Ablation Study}
\label{sec:experiment_ablation}

\begin{figure}[H]
    \centering
    \includegraphics[width=0.98\textwidth]{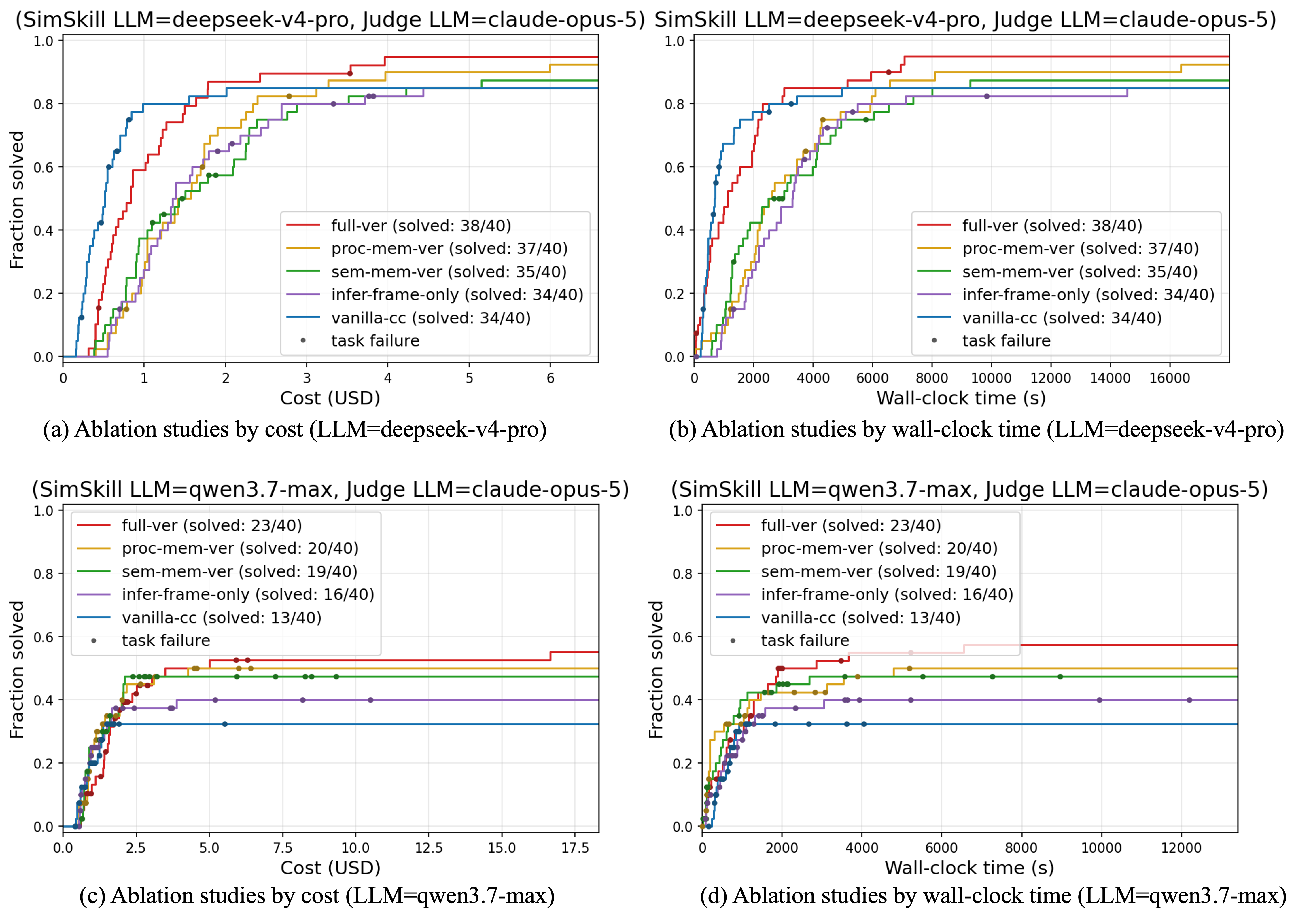}
    \caption{Benchmark V1 ablation curves for DeepSeek-V4-Pro (top) and Qwen3.7-Max (bottom), under the Claude Opus~5 binary judge. The left panels use dollar cost and the right panels use wall-clock time.}
    \label{fig:ablations}
\end{figure}

Figure~\ref{fig:ablations} and Table~\ref{tab:ablation_endpoints} compare all five conditions on V1. For DeepSeek-V4-Pro, the inference framework without memory solves the same 34 tasks as vanilla Claude Code. Adding semantic memory raises the endpoint to 35, procedural memory raises it to 37, and the combination reaches 38. For Qwen3.7-Max, the framework itself raises completion from 13 to 16 tasks; semantic and procedural memory raise it further to 19 and 20; and the complete system reaches 23. Procedural memory contributes slightly more than semantic memory for both backbones, but neither representation subsumes the other.

\begin{table*}[t]
    \centering
    \small
    \setlength{\tabcolsep}{4pt}
    \caption{Verified V1 endpoint in the five-way ablation (tasks solved out of 40).}
    \label{tab:ablation_endpoints}
    \begin{tabular}{lccccc}
        \hline
        Backbone & \texttt{vanilla-cc} & \texttt{infer-frame-only} & \texttt{sem-mem-ver} & \texttt{proc-mem-ver} & \texttt{full-ver} \\
        \hline
        DeepSeek-V4-Pro & 34 & 34 & 35 & 37 & 38 \\
        Qwen3.7-Max     & 13 & 16 & 19 & 20 & 23 \\
        \hline
    \end{tabular}
\end{table*}

The factorial interaction contrast at the completion endpoint is
\begin{equation}
    \Delta_{P\times S}
    = A_{PS}-A_{P\bar{S}}-A_{\bar{P}S}+A_{\bar{P}\bar{S}},
    \label{eq:memory_interaction}
\end{equation}
where $P$ and $S$ denote the presence of procedural and semantic memory within the SimSkill inference framework. Equation~\ref{eq:memory_interaction} equals zero for both backbones: $0.950-0.925-0.875+0.850=0$ for DeepSeek-V4-Pro and $0.575-0.500-0.475+0.400=0$ for Qwen3.7-Max. At the aggregate endpoint, the two memory types therefore exhibit additive rather than super-additive effects in this ablation study.

The resource curves add an important observation. With DeepSeek-V4-Pro, the complete system has a lower median cost and time than either single-memory ablation despite solving more tasks: \$0.78 and 1056~s, compared with \$1.41 and 2423~s for procedural memory alone and \$1.31 and 2383~s for semantic memory alone. Access to both forms of memory appears to reduce unproductive exploration, even though their endpoint accuracy effects are additive. For Qwen3.7-Max, by contrast, the complete system is more expensive and slower than either single-memory variant, and its advantage emerges mainly in the long-budget tail. This is again a backbone-dependent result, and it shows that the value of an architectural component must be stated jointly with the operating budget. In terms of final success the ordering is \texttt{full-ver} $>$ \texttt{proc-mem-ver} $>$ \texttt{sem-mem-ver} $>$ \texttt{infer-frame-only} $\geq$ \texttt{vanilla-cc}, but it does not hold at every intermediate cost or time threshold; the final inequality would be an equality for the stronger backbone.

\subsection{Scope of the Evidence}
\label{sec:experiment_scope}

Taken together, the results answer the five research questions with important qualifications. Full SimSkill increases verified success for two of the three tested backbones and retains an advantage on hard compositional and novel tasks, but the effect depends on the backbone model. Both procedural and semantic memory contribute, with a modestly larger procedural contribution. Memory can reduce wasted exploration in some settings, yet its retrieval and orchestration overhead frequently raises total inference cost; the principal demonstrated benefit is expanded task coverage rather than universal cost savings.

\section{Conclusion}
\label{sec:conclusion}

This paper presented SimSkill, a self-evolving LLM agent that converts interaction with an executable environment into persistent competence without updating the backbone model. In SUMO, autonomous curriculum generation and task execution are coupled with episodic, procedural, and semantic memory. Procedural memory is stored as skills and semantic memory as linked knowledge pages following the LLM Wiki pattern, while memory ingestion and linting support their continuous accumulation and maintenance. Both the control logic and the stored artifacts are written primarily in natural language, with executable scripts bundled into skills where operations must be repeated exactly. After approximately 80 hours of operation, SimSkill had accumulated 150 procedural skills and 153 semantic-memory pages, all in open text formats that other models and agent frameworks can read and reuse.

On two held-out 40-task benchmarks judged independently, SimSkill improved verified success for DeepSeek-V4-Pro and Qwen3.7-Max, with gains concentrated in the most demanding tasks---evidence that accumulated experience is recombined rather than merely replayed. The ablations indicate complementary, additive contributions from procedural and semantic memory. The gains were not universal: GLM-5.2 showed no improvement, and greater task coverage did not consistently reduce inference cost. The effect depends on the backbone model.

Future work should evaluate successive memory snapshots, and improve curriculum selection, experience consolidation, and retrieval at larger memory scales. Transfer and forgetting should also be tested across simulators and scientific domains. Explicit memory may also be combined with parameter adaptation.

More broadly, SimSkill exemplifies a shift from fully hard-coded agent workflows toward systems whose goals, principles, knowledge, and procedures are expressed in natural language and composed by an LLM with executable tools. This division parallels collective human problem solving: language preserves and transmits adaptable knowledge, while tools, algorithms, and code provide precise and reproducible execution. Natural language can therefore serve not only as an interface, but also as a medium for accumulating and reorganizing computational capability. We expect this way of building systems to spread well beyond traffic simulation: as backbone models improve, a growing share of what is now written as code will instead be written as language, changing how software systems are designed, maintained, and shared.

\bibliographystyle{plainnat}
\bibliography{main}

@inproceedings{pertsch2021accelerating,
  title={Accelerating reinforcement learning with learned skill priors},
  author={Pertsch, Karl and Lee, Youngwoon and Lim, Joseph},
  booktitle={Conference on robot learning},
  pages={188--204},
  year={2021},
  organization={PMLR}
}

@article{wang2024comprehensive,
  title={A comprehensive survey of continual learning: Theory, method and application},
  author={Wang, Liyuan and Zhang, Xingxing and Su, Hang and Zhu, Jun},
  journal={IEEE transactions on pattern analysis and machine intelligence},
  volume={46},
  number={8},
  pages={5362--5383},
  year={2024},
  publisher={IEEE}
}

@book{tulving2000oxford,
  title={The Oxford handbook of memory},
  author={Tulving, Endel and Craik, Fergus IM},
  year={2000},
  publisher={Oxford University Press}
}

@article{squire2004memory,
  title={Memory systems of the brain: A brief history and current perspective},
  author={Squire, Larry R},
  journal={Neurobiology of learning and memory},
  volume={82},
  number={3},
  pages={171--177},
  year={2004},
  publisher={Elsevier}
}

@article{sutton1999options,
  title={Between {MDPs} and Semi-{MDPs}: A Framework for Temporal Abstraction in Reinforcement Learning},
  author={Sutton, Richard S. and Precup, Doina and Singh, Satinder},
  journal={Artificial Intelligence},
  volume={112},
  number={1--2},
  pages={181--211},
  year={1999},
  doi={10.1016/S0004-3702(99)00052-1}
}

@misc{sutton2019bitterlesson,
  author={Sutton, Richard S.},
  title={The Bitter Lesson},
  year={2019},
  howpublished={\url{http://www.incompleteideas.net/IncIdeas/BitterLesson.html}},
  note={Incomplete Ideas}
}

@misc{anthropic2026skills,
  author={{Anthropic}},
  title={Extend Claude with Skills},
  year={2026},
  howpublished={\url{https://code.claude.com/docs/en/skills}},
  note={Claude Code documentation. Accessed: 2026-08-24}
}

@article{da2024openti,
  title={Open-ti: Open traffic intelligence with augmented language model},
  author={Da, Longchao and Liou, Kuanru and Chen, Tiejin and Zhou, Xuesong and Luo, Xiangyong and Yang, Yezhou and Wei, Hua},
  journal={International Journal of Machine Learning and Cybernetics},
  volume={15},
  number={10},
  pages={4761--4786},
  year={2024},
  publisher={Springer}
}

@article{li2026chatsumoagent,
  title={ChatSUMO Agent: An LLM-Based Agent for Conversational Traffic Simulation in SUMO},
  author={Li, Shuyang and Ma, Meng and Azfar, Talha and Ke, Ruimin},
  journal={Transportation Research Part C: Emerging Technologies},
  volume={190},
  pages={105759},
  year={2026},
  publisher={Elsevier}
}

@misc{sumo2026,
  author = {{Eclipse Foundation}},
  title = {{Eclipse SUMO -- Simulation of Urban MObility}},
  year = {2026},
  howpublished = {\url{https://eclipse.dev/sumo/}},
  note = {Accessed: 2026-02-14}
}

@inproceedings{colas2023augmenting,
  title={Augmenting autotelic agents with large language models},
  author={Colas, C{\'e}dric and Teodorescu, Laetitia and Oudeyer, Pierre-Yves and Yuan, Xingdi and C{\^o}t{\'e}, Marc-Alexandre},
  booktitle={Conference on Lifelong Learning Agents},
  pages={205--226},
  year={2023},
  organization={PMLR}
}

@inproceedings{zhang2025appagent,
  title={Appagent: Multimodal agents as smartphone users},
  author={Zhang, Chi and Yang, Zhao and Liu, Jiaxuan and Li, Yanda and Han, Yucheng and Chen, Xin and Huang, Zebiao and Fu, Bin and Yu, Gang},
  booktitle={Proceedings of the 2025 CHI Conference on Human Factors in Computing Systems},
  pages={1--20},
  year={2025}
}

@article{wang2024survey,
  title={A survey on large language model based autonomous agents},
  author={Wang, Lei and Ma, Chen and Feng, Xueyang and Zhang, Zeyu and Yang, Hao and Zhang, Jingsen and Chen, Zhiyuan and Tang, Jiakai and Chen, Xu and Lin, Yankai and others},
  journal={Frontiers of Computer Science},
  volume={18},
  number={6},
  pages={186345},
  year={2024},
  publisher={Springer}
}

@misc{multica2026karpathyskills,
  author       = {{Multica AI}},
  title        = {{andrej-karpathy-skills}: A single {CLAUDE}.md file to improve Claude Code behavior, derived from Andrej Karpathy's observations on LLM coding pitfalls},
  year         = {2026},
  howpublished = {\url{https://github.com/multica-ai/andrej-karpathy-skills}},
  note         = {GitHub repository. Accessed: 2026-06-26}
}

@inproceedings{haddouch2018modeling,
  title={Modeling the flow of road traffic with the SUMO simulator},
  author={Haddouch, Sara and Hachimi, Hana{\^a} and Hmina, Nabil},
  booktitle={2018 4th International Conference on Optimization and Applications (ICOA)},
  pages={1--5},
  year={2018},
  organization={IEEE}
}

@inproceedings{ejercito2017traffic,
  title={Traffic simulation software review},
  author={Ejercito, Paolo M and Nebrija, Kristine Gayle E and Feria, Rommel P and Lara-Figueroa, Ligaya Leah},
  booktitle={2017 8th International Conference on Information, Intelligence, Systems \& Applications (IISA)},
  pages={1--4},
  year={2017},
  organization={IEEE}
}

@article{gulcehre2023reinforced,
  title={Reinforced self-training (rest) for language modeling},
  author={Gulcehre, Caglar and Paine, Tom Le and Srinivasan, Srivatsan and Konyushkova, Ksenia and Weerts, Lotte and Sharma, Abhishek and Siddhant, Aditya and Ahern, Alex and Wang, Miaosen and Gu, Chenjie and others},
  journal={arXiv preprint arXiv:2308.08998},
  year={2023}
}

@inproceedings{wang2023self,
  title={Self-instruct: Aligning language models with self-generated instructions},
  author={Wang, Yizhong and Kordi, Yeganeh and Mishra, Swaroop and Liu, Alisa and Smith, Noah A and Khashabi, Daniel and Hajishirzi, Hannaneh},
  booktitle={Proceedings of the 61st annual meeting of the association for computational linguistics (volume 1: long papers)},
  pages={13484--13508},
  year={2023}
}

@article{tao2024survey,
  title={A survey on self-evolution of large language models},
  author={Tao, Zhengwei and Lin, Ting-En and Chen, Xiancai and Li, Hangyu and Wu, Yuchuan and Li, Yongbin and Jin, Zhi and Huang, Fei and Tao, Dacheng and Zhou, Jingren},
  journal={arXiv preprint arXiv:2404.14387},
  year={2024}
}

@article{shinn2023reflexion,
  title={Reflexion: Language agents with verbal reinforcement learning},
  author={Shinn, Noah and Cassano, Federico and Gopinath, Ashwin and Narasimhan, Karthik and Yao, Shunyu},
  journal={Advances in neural information processing systems},
  volume={36},
  pages={8634--8652},
  year={2023}
}

@article{yao2022react,
  title={React: Synergizing reasoning and acting in language models},
  author={Yao, Shunyu and Zhao, Jeffrey and Yu, Dian and Du, Nan and Shafran, Izhak and Narasimhan, Karthik and Cao, Yuan},
  journal={arXiv preprint arXiv:2210.03629},
  year={2022}
}

@article{yang2026skillopt,
  title={Skillopt: Executive strategy for self-evolving agent skills},
  author={Yang, Yifan and Gong, Ziyang and Huang, Weiquan and Yang, Qihao and Zhou, Ziwei and Huang, Zisu and Li, Yan and Gao, Xuemei and Dai, Qi and Liu, Bei and others},
  journal={arXiv preprint arXiv:2605.23904},
  year={2026}
}

@article{zheng2026lifelong,
  title={Lifelong learning of large language model based agents: A roadmap},
  author={Zheng, Junhao and Shi, Chengming and Cai, Xidi and Li, Qiuke and Zhang, Duzhen and Li, Chenxing and Yu, Dong and Ma, Qianli},
  journal={IEEE Transactions on Pattern Analysis and Machine Intelligence},
  year={2026},
  publisher={IEEE}
}

@article{zhang2025survey,
  title={A survey on the memory mechanism of large language model-based agents},
  author={Zhang, Zeyu and Dai, Quanyu and Bo, Xiaohe and Ma, Chen and Li, Rui and Chen, Xu and Zhu, Jieming and Dong, Zhenhua and Wen, Ji-Rong},
  journal={ACM Transactions on Information Systems},
  volume={43},
  number={6},
  pages={1--47},
  year={2025},
  publisher={ACM New York, NY}
}

@inproceedings{salama2025meminsight,
  title={Meminsight: Autonomous memory augmentation for llm agents},
  author={Salama, Rana and Cai, Jason and Yuan, Michelle and Currey, Anna and Sunkara, Monica and Zhang, Yi and Benajiba, Yassine},
  booktitle={Proceedings of the 2025 Conference on Empirical Methods in Natural Language Processing},
  pages={33124--33140},
  year={2025}
}

@article{xu2026mem,
  title={A-mem: Agentic memory for llm agents},
  author={Xu, Wujiang and Liang, Zujie and Mei, Kai and Gao, Hang and Tan, Juntao and Zhang, Yongfeng},
  journal={Advances in Neural Information Processing Systems},
  volume={38},
  pages={17577--17604},
  year={2026}
}

@article{chhikara2025mem0,
  title={Mem0: Building production-ready ai agents with scalable long-term memory},
  author={Chhikara, Prateek and Khant, Dev and Aryan, Saket and Singh, Taranjeet and Yadav, Deshraj},
  journal={arXiv preprint arXiv:2504.19413},
  year={2025}
}

@article{liu2025generative,
  title={Generative agents for urban mobility: A cognitive framework for realistic travel behavior simulation},
  author={Liu, Qi and Li, Can and Ma, Wanjing},
  journal={Simulation Modelling Practice and Theory},
  pages={103234},
  year={2025},
  publisher={Elsevier}
}

@article{liu2026gatsim,
  title={Gatsim: Urban mobility simulation with generative agents},
  author={Liu, Qi and Li, Can and Ma, Wanjing},
  journal={Transportation Research Part C: Emerging Technologies},
  volume={186},
  pages={105576},
  year={2026},
  publisher={Elsevier}
}

@inproceedings{park2023generative,
  title={Generative agents: Interactive simulacra of human behavior},
  author={Park, Joon Sung and O'Brien, Joseph and Cai, Carrie Jun and Morris, Meredith Ringel and Liang, Percy and Bernstein, Michael S},
  booktitle={Proceedings of the 36th annual acm symposium on user interface software and technology},
  pages={1--22},
  year={2023}
}

@inproceedings{zhong2024memorybank,
  title={Memorybank: Enhancing large language models with long-term memory},
  author={Zhong, Wanjun and Guo, Lianghong and Gao, Qiqi and Ye, He and Wang, Yanlin},
  booktitle={Proceedings of the AAAI conference on artificial intelligence},
  pages={19724--19731},
  year={2024}
}

@article{packer2023memgpt,
  title={Memgpt: Towards llms as operating systems},
  author={Packer, Charles and Wooders, Sarah and Lin, Kevin and Fang, Vivian and Patil, Shishir G and Stoica, Ion and Gonzalez, Joseph E},
  journal={arXiv preprint arXiv:2310.08560},
  year={2023}
}

@article{lewis2020retrieval,
  title={Retrieval-augmented generation for knowledge-intensive nlp tasks},
  author={Lewis, Patrick and Perez, Ethan and Piktus, Aleksandra and Petroni, Fabio and Karpukhin, Vladimir and Goyal, Naman and K{\"u}ttler, Heinrich and Lewis, Mike and Yih, Wen-tau and Rockt{\"a}schel, Tim and others},
  journal={Advances in neural information processing systems},
  volume={33},
  pages={9459--9474},
  year={2020}
}

@article{qu2024recursive,
  title={Recursive introspection: Teaching language model agents how to self-improve},
  author={Qu, Yuxiao and Zhang, Tianjun and Garg, Naman and Kumar, Aviral},
  journal={Advances in Neural Information Processing Systems},
  volume={37},
  pages={55249--55285},
  year={2024}
}

@misc{mcveety2026okf,
  author       = {Sam McVeety and Amir Hormati},
  title        = {How the Open Knowledge Format Can Improve Data Sharing},
  year         = {2026},
  month        = jun,
  howpublished = {\url{https://cloud.google.com/blog/products/data-analytics/how-the-open-knowledge-format-can-improve-data-sharing}},
  note         = {Google Cloud Blog, accessed June 23, 2026}
}

@article{chen2024data,
  title={Data-driven traffic simulation: A comprehensive review},
  author={Chen, Di and Zhu, Meixin and Yang, Hao and Wang, Xuesong and Wang, Yinhai},
  journal={IEEE Transactions on Intelligent Vehicles},
  volume={9},
  number={4},
  pages={4730--4748},
  year={2024},
  publisher={IEEE}
}

@inproceedings{huang2023large,
  title={Large language models can self-improve},
  author={Huang, Jiaxin and Gu, Shixiang and Hou, Le and Wu, Yuexin and Wang, Xuezhi and Yu, Hongkun and Han, Jiawei},
  booktitle={Proceedings of the 2023 conference on empirical methods in natural language processing},
  pages={1051--1068},
  year={2023}
}

@article{wang2023voyager,
  title={Voyager: An open-ended embodied agent with large language models},
  author={Wang, Guanzhi and Xie, Yuqi and Jiang, Yunfan and Mandlekar, Ajay and Xiao, Chaowei and Zhu, Yuke and Fan, Linxi and Anandkumar, Anima},
  journal={arXiv preprint arXiv:2305.16291},
  year={2023}
}

@misc{karpathy2026wiki,
  author       = {Karpathy, Andrej},
  title        = {{LLM Wiki}},
  howpublished = {\url{https://gist.github.com/karpathy/442a6bf555914893e9891c11519de94f}},
  year         = {2026}
}

@article{ye2025sumo,
  title={SUMO-MCP: Leveraging the model context protocol for autonomous traffic simulation and optimization},
  author={Ye, Chenglong and Xiong, Gang and Shang, Junyou and Dai, Xingyuan and Gong, Xiaoyan and Lv, Yisheng},
  journal={arXiv preprint arXiv:2506.03548},
  year={2025}
}

@article{li2024chatsumo,
  title={Chatsumo: Large language model for automating traffic scenario generation in simulation of urban mobility},
  author={Li, Shuyang and Azfar, Talha and Ke, Ruimin},
  journal={IEEE Transactions on Intelligent Vehicles},
  year={2024},
  publisher={IEEE}
}

@article{zhu2023ghost,
  title={Ghost in the minecraft: Generally capable agents for open-world environments via large language models with text-based knowledge and memory},
  author={Zhu, Xizhou and Chen, Yuntao and Tian, Hao and Tao, Chenxin and Su, Weijie and Yang, Chenyu and Huang, Gao and Li, Bin and Lu, Lewei and Wang, Xiaogang and others},
  journal={arXiv preprint arXiv:2305.17144},
  year={2023}
}

@article{wang2024jarvis,
  title={Jarvis-1: Open-world multi-task agents with memory-augmented multimodal language models},
  author={Wang, Zihao and Cai, Shaofei and Liu, Anji and Jin, Yonggang and Hou, Jinbing and Zhang, Bowei and Lin, Haowei and He, Zhaofeng and Zheng, Zilong and Yang, Yaodong and others},
  journal={IEEE Transactions on Pattern Analysis and Machine Intelligence},
  volume={47},
  number={3},
  pages={1894--1907},
  year={2024},
  publisher={IEEE}
}

@article{zhang2024trafficgpt,
  title={Trafficgpt: Viewing, processing and interacting with traffic foundation models},
  author={Zhang, Siyao and Fu, Daocheng and Liang, Wenzhe and Zhang, Zhao and Yu, Bin and Cai, Pinlong and Yao, Baozhen},
  journal={Transport Policy},
  volume={150},
  pages={95--105},
  year={2024},
  publisher={Elsevier}
}

@inproceedings{lai2025llmlight,
  title={LLMLight: Large language models as traffic signal control agents},
  author={Lai, Siqi and Xu, Zhao and Zhang, Weijia and Liu, Hao and Xiong, Hui},
  booktitle={Proceedings of the 31st ACM SIGKDD Conference on Knowledge Discovery and Data Mining V. 1},
  pages={2335--2346},
  year={2025}
}

@article{wang2024llm,
  title={LLM-assisted light: Leveraging large language model capabilities for human-mimetic traffic signal control in complex urban environments},
  author={Wang, Maonan and Pang, Aoyu and Kan, Yuheng and Pun, Man-On and Chen, Chung Shue and Huang, Bo},
  journal={arXiv preprint arXiv:2403.08337},
  year={2024}
}

@article{du2025trafficsimagent,
  title={TrafficSimAgent: A Hierarchical Agent Framework for Autonomous Traffic Simulation with MCP Control},
  author={Du, Yuwei and Zhang, Jun and Feng, Jie and Liu, Zhicheng and Yuan, Jian and Li, Yong},
  journal={arXiv preprint arXiv:2512.20996},
  year={2025}
}

@inproceedings{tziafas2024lifelong,
  title={Lifelong robot library learning: Bootstrapping composable and generalizable skills for embodied control with language models},
  author={Tziafas, Georgios and Kasaei, Hamidreza},
  booktitle={2024 IEEE International Conference on Robotics and Automation (ICRA)},
  pages={515--522},
  year={2024},
  organization={IEEE}
}

@article{yang2026autoskill,
  title={Autoskill: Experience-driven lifelong learning via skill self-evolution},
  author={Yang, Yutao and Li, Junsong and Pan, Qianjun and Zhan, Bihao and Cai, Yuxuan and Du, Lin and Zhou, Jie and Chen, Kai and Chen, Qin and Li, Xin and others},
  journal={arXiv preprint arXiv:2603.01145},
  year={2026}
}

@article{zheng2025lifelongagentbench,
  title={Lifelongagentbench: Evaluating llm agents as lifelong learners},
  author={Zheng, Junhao and Cai, Xidi and Li, Qiuke and Zhang, Duzhen and Li, ZhongZhi and Zhang, Yingying and Song, Le and Ma, Qianli},
  journal={arXiv preprint arXiv:2505.11942},
  year={2025}
}

@article{jeong2025agentsumo,
  title={AgentSUMO: An Agentic Framework for Interactive Simulation Scenario Generation in SUMO via Large Language Models},
  author={Jeong, Minwoo and Chang, Jeeyun and Yoon, Yoonjin},
  journal={arXiv preprint arXiv:2511.06804},
  year={2025}
}

@inproceedings{behrisch2011sumo,
  title={SUMO--simulation of urban mobility: an overview},
  author={Behrisch, Michael and Bieker, Laura and Erdmann, Jakob and Krajzewicz, Daniel},
  booktitle={Proceedings of SIMUL 2011, the third international conference on advances in system simulation},
  year={2011},
  organization={ThinkMind}
}

@book{sutton1998reinforcement,
  title={Reinforcement learning: An introduction},
  author={Sutton, Richard S and Barto, Andrew G and others},
  year={1998},
  publisher={MIT press Cambridge}
}

@article{kirkpatrick2017overcoming,
  title={Overcoming catastrophic forgetting in neural networks},
  author={Kirkpatrick, James and Pascanu, Razvan and Rabinowitz, Neil and Veness, Joel and Desjardins, Guillaume and Rusu, Andrei A and Milan, Kieran and Quan, John and Ramalho, Tiago and Grabska-Barwinska, Agnieszka and others},
  journal={Proceedings of the national academy of sciences},
  volume={114},
  number={13},
  pages={3521--3526},
  year={2017},
  publisher={National Academy of Sciences}
}

\clearpage
\appendix

\section{Complete Categorized Memory Inventory}
\label{app:skill_inventory}

Listing~\ref{lst:complete_skill_inventory} gives the complete snapshot summarized in Table~\ref{tab:skill_coverage}. Knowledge pages and procedural skills are independently alphabetized within each category; entries appearing on the same row are not necessarily paired. Categories indicate primary function only, as cross-links and composition routinely connect artifacts across these boundaries.

\begin{lstlisting}[caption={Complete categorized inventory of the 153 semantic knowledge pages and 150 procedural skills.},
                   label={lst:complete_skill_inventory},
                   basicstyle=\ttfamily\scriptsize,
                   breaklines=false,
                   columns=fixed,
                   keepspaces=true]
[Scenario construction, execution, and vehicle-state operations]
Knowledge pages (6)                          | Procedural skills (6)
---------------------------------------------+---------------------------------------------
* change-vehicle-state                       | * analyze-simulation-outputs
* mesoscopic-simulation                      | * choose-time-discretization-and-
                                             |   integration-method
* sumo-command-line                          | * get-vehicles-state
* sumo-output-files                          | * run-mesoscopic-simulation
* sumo-time-discretization                   | * run-simulation
* traci                                      | * set-vehicle-state

[Network and infrastructure design]
Knowledge pages (11)                         | Procedural skills (12)
---------------------------------------------+---------------------------------------------
* abstract-network-generation                | * audit-repair-and-persist-imported-network-
                                             |   defects
* cutroutes-and-subnetwork-extraction        | * compare-one-way-vs-two-way-street-grid-
                                             |   conversion
* horizontal-curvature-and-curve-speed-in-   | * create-grid-network
  sumo                                       |
* imported-network-defect-classes-and-       | * create-roundabout-network
  traffic-impact                             |
* multi-resolution-modeling-buffer-sizing-   | * create-single-intersection
  and-boundary-handoff                       |
* one-way-vs-two-way-grid-performance-       | * create-spider-network
  crossover                                  |
* opendrive-and-network-format-              | * extract-subnetwork-scenario-with-boundary-
  interoperability                           |   demand
* openstreetmap                              | * load-osm-network
* road-gradient-and-energy-consumption       | * model-horizontal-curvature-and-evaluate-
                                             |   design-consistency
* roundabout-modeling-and-comparison         | * model-road-gradient-effects-on-energy
* vehicle-class-lane-permissions             | * model-vclass-lane-permissions
                                             | * quantify-opendrive-roundtrip-fidelity

[Demand, routing, and assignment]
Knowledge pages (17)                         | Procedural skills (17)
---------------------------------------------+---------------------------------------------
* activitygen                                | * assign-traffic-with-marouter
* braess-paradox-in-sumo                     | * build-four-step-model-with-feedback-loop
* dfrouter-detector-based-demand-            | * compute-dynamic-user-equilibrium
  reconstruction                             |
* downs-thomson-paradox-and-mode-choice-     | * construct-and-verify-braess-paradox
  equilibrium                                |
* duarouter                                  | * convert-od-matrix-to-trips
* dynamic-user-equilibrium-and-wardrop       | * convert-trips-to-routes
* effort-based-routing-and-eco-routing       | * equilibrate-departure-time-choice-in-
                                             |   bottleneck-model
* field-counts-to-simulation-demand-and-the- | * equilibrate-endogenous-mode-choice-with-
  saturated-count-truncation-trap            |   transit-supply-feedback
* four-step-model-feedback-loop-convergence  | * generate-activity-based-demand
* gps-map-matching-and-probe-demand-         | * generate-demand-with-jtrrouter
  reconstruction                             |
* jtrrouter                                  | * generate-random-trips
* marouter-macroscopic-assignment            | * implement-eco-routing
* od2trips                                   | * map-match-gps-traces-to-reconstruct-demand
* population-synthesis-and-aggregation-bias  | * reconstruct-demand-with-dfrouter
* random-trips                               | * reconstruct-simulation-demand-from-field-
                                             |   turning-movement-counts
* route-choice-model-verification-overlap-   | * specify-route-choice-models-and-generate-
  and-route-set-effects                      |   route-sets
* vickrey-bottleneck-departure-time-         | * synthesize-population-and-generate-
  equilibrium                                |   disaggregate-demand

[Signals and intersection control]
Knowledge pages (31)                         | Procedural skills (30)
---------------------------------------------+---------------------------------------------
* actuated-signal-detector-design-and-fault- | * build-atspm-pipeline-and-retime-arterial
  tolerance                                  |
* actuated-traffic-signals                   | * build-pedestrian-crossings-and-phasing
* arterial-signal-progression-resonance-     | * compare-left-turn-signal-treatments
  bandwidth-and-delay                        |
* automated-traffic-signal-performance-      | * compare-unsignalized-intersection-control-
  measures                                   |   types
* autonomous-intersection-management-safety- | * conduct-driveway-signal-warrant-traffic-
  and-performance-envelope                   |   impact-analysis
* connected-vehicle-penetration-and-         | * control-signals-with-actuated-tls
  detector-free-signal-control               |
* coordinated-adaptive-signal-control-       | * design-actuated-signal-detector-placement-
  detector-bias-and-transition-cost          |   and-fault-tolerance
* emergency-vehicle-preemption-and-bluelight | * design-arterial-signal-progression-and-
                                             |   verify-bandwidth
* glosa-eco-driving                          | * design-left-turn-storage-bay-length
* intersection-sight-distance-and-sumo-      | * design-multimodal-signal-progression-for-
  visibility-parameter                       |   bicycles-and-cars
* left-turn-storage-bay-length-design        | * design-restricted-crossing-uturn-and-
                                             |   michigan-left-intersections
* left-turn-treatment-tradeoffs              | * design-signal-change-and-clearance-
                                             |   intervals
* max-pressure-signal-control                | * evaluate-right-turn-on-red-and-leading-
                                             |   pedestrian-interval
* multimodal-signal-progression-and-the-     | * implement-detector-free-cv-adaptive-
  bicycle-green-wave                         |   signal-control
* mutcd-signal-warrants-and-the-demand-vs-   | * implement-emergency-vehicle-preemption
  served-volume-trap                         |
* nema-dual-ring-controller                  | * implement-glosa-speed-advisory-controller
* pedestrian-crossings-and-signal-phasing    | * implement-maxpressure-traci-controller
* q-learning-agent                           | * implement-nema-dual-ring-controller
* railroad-preemption-of-nearby-signalized-  | * implement-predictive-rolling-horizon-
  intersections                              |   signal-control
* rcut-and-michigan-left-alternative-        | * implement-railroad-preemption-at-a-
  intersection-design                        |   signalized-intersection
* right-turn-on-red-and-leading-pedestrian-  | * implement-reservation-based-autonomous-
  interval                                   |   intersection-management
* roundabout-capacity-law-and-demand-        | * implement-scats-style-coordinated-
  metering                                   |   adaptive-signal-control
* signal-clearance-intervals-dilemma-zone-   | * implement-transit-signal-priority
  and-safety-capacity-tradeoff               |
* simulation-in-the-loop-ga-signal-          | * measure-roundabout-capacity-and-implement-
  optimization                               |   metering
* sumo-rl-environment                        | * model-intersection-sight-distance-
                                             |   restriction-at-a-twsc-junction
* tlscoordinator                             | * optimize-signal-plan-with-simulation-in-
                                             |   the-loop-ga
* tlscycleadaptation                         | * optimize-signals-by-qlearning
* transit-signal-priority                    | * optimize-signals-by-tlscoordinator
* unsignalized-vs-signalized-intersection-   | * optimize-signals-by-tlscycleadaptation
  control                                    |
* value-of-anticipation-in-predictive-       | * switch-signal-plans-by-time-of-day-with-
  signal-control                             |   waut
* waut-time-of-day-signal-plan-switching     |

[Freeway, corridor, and network operations]
Knowledge pages (31)                         | Procedural skills (31)
---------------------------------------------+---------------------------------------------
* automatic-incident-detection-algorithms    | * build-and-benchmark-freeway-incident-
                                             |   detection
* coordinated-ramp-metering-delay-transfer-  | * build-and-evaluate-system-interchange
  and-ramp-storage                           |
* cordon-tolling-and-e3-detectors            | * build-diamond-interchange-with-signal-
                                             |   offset-spillback
* corridor-access-management-twltl-          | * build-diverging-diamond-interchange
  representation-and-density-effects         |
* diamond-interchange-signal-offset-and-     | * compare-zipper-vs-default-merge-at-lane-
  spillback                                  |   drop
* discrete-network-design-and-project-       | * control-one-lane-two-way-alternating-flow-
  interaction                                |   through-a-work-zone
* diverging-diamond-interchange-unopposed-   | * demonstrate-and-stabilize-phantom-traffic-
  lefts                                      |   jams
* dynamic-hard-shoulder-running-with-traci-  | * design-and-control-freeway-work-zone-lane-
  lane-permissions                           |   closures
* evacuation-clearance-time-analysis         | * evaluate-corridor-access-management-and-
                                             |   median-treatments
* freeway-weaving-segment-turbulence         | * evaluate-integrated-corridor-management-
                                             |   with-factorial-interaction-design
* freeway-work-zone-capacity-closure-        | * evaluate-neighborhood-traffic-calming-and-
  representation-and-merge-control           |   cut-through-displacement
* grade-aware-heavy-vehicle-physics-and-     | * evaluate-two-lane-highway-with-hcm-and-
  climbing-lane-warrants                     |   passing-lanes
* incident-rerouting-and-closures            | * form-platoons-with-simpla
* information-penetration-and-congestible-   | * implement-alinea-ramp-metering
  routing                                    |
* integrated-corridor-management-factorial-  | * implement-coordinated-corridor-ramp-
  interaction-findings                       |   metering
* managed-lanes-empty-lane-paradox-and-      | * implement-dynamic-hard-shoulder-running
  person-throughput                          |
* mfd-based-perimeter-gating                 | * implement-mfd-based-perimeter-gating
* neighborhood-traffic-calming-displacement- | * implement-variable-speed-limits
  and-evaporation                            |
* network-link-criticality-and-proxy-        | * model-adverse-weather-effects-on-freeway-
  validation                                 |   traffic
* one-lane-two-way-alternating-flow-and-     | * model-cordon-tolling-with-generalized-
  shared-lane-representation                 |   cost-surcharge
* opposite-direction-overtaking-mechanics    | * model-freeway-weaving-segment
* phantom-traffic-jams-and-single-av-        | * model-grade-aware-heavy-vehicle-
  stabilization                              |   performance-and-climbing-lanes
* ramp-metering-with-alinea                  | * model-managed-lanes-with-dynamic-tolling-
                                             |   and-self-selection
* reversible-lane-encoding-and-changeover-   | * model-opposite-direction-overtaking
  safety                                     |
* simpla-platooning                          | * model-toll-plaza-as-queueing-facility
* system-interchange-weaving-and-design-     | * operate-reversible-tidal-flow-lane
  selection                                  |
* toll-plaza-queueing-and-the-service-       | * scan-network-link-criticality-and-
  headway-floor                              |   vulnerability
* two-lane-highway-follower-density-and-     | * simulate-emergency-evacuation
  passing-lane-effectiveness                 |
* variable-speed-limits-and-e2-detectors     | * simulate-incident-rerouting
* weather-friction-effects-on-capacity-and-  | * solve-budget-constrained-network-design-
  safety                                     |   problem
* zipper-merge-lane-drop-discharge           | * sweep-rerouting-device-market-penetration

[Transit, multimodal, fleet, and parking systems]
Knowledge pages (22)                         | Procedural skills (20)
---------------------------------------------+---------------------------------------------
* battery-electric-bus-energy-and-charger-   | * build-and-evaluate-park-and-ride-corridor
  sizing                                     |
* bus-bunching-and-forward-headway-holding   | * build-gtfs-transit-scenario
* bus-stop-infrastructure-design-parking-    | * build-rail-corridor-with-railsignal
  mechanism-and-tsp-interaction              |
* car-to-transit-intermodal-transfer-and-    | * build-rail-road-grade-crossing
  park-and-ride                              |
* cruising-for-parking-search-externality-   | * demonstrate-and-control-bus-bunching
  and-remedies                               |
* curbside-delivery-blocking-externality     | * design-bus-stop-placement-type-and-spacing
* dedicated-bicycle-lanes-and-mode-share     | * design-transit-service-plan-under-a-bus-
                                             |   hour-budget
* electric-vehicle-battery-and-charging      | * evaluate-protected-bicycle-intersection-
                                             |   design
* gtfs-import-and-pt-representation-         | * model-capacity-constrained-transit-
  semantics                                  |   passenger-loading
* intermodal-transfer-and-person-stage-      | * model-cruising-for-parking-search-
  semantics-in-sumo                          |   externality
* parking-areas-and-rerouters                | * model-curbside-delivery-and-lane-blocking-
                                             |   externality
* protected-bicycle-intersection-design-and- | * model-dedicated-bicycle-lane-
  right-hook-mechanics                       |   infrastructure
* public-transport-and-intermodal-routing    | * model-parking-with-rerouting
* rail-crossing-junction-mechanics           | * model-urban-freight-delivery-tours
* rail-simulation-and-railsignal             | * simulate-ev-charging
* station-based-shared-micromobility-in-sumo | * simulate-motorcycle-lane-filtering-with-
                                             |   sublane-model
* street-running-tram-reservation-and-right- | * simulate-multimodal-transit
  of-way-tradeoffs                           |
* sublane-model-and-lane-filtering           | * simulate-street-running-tram-corridor
* taxi-and-drt-dispatch                      | * simulate-taxi-and-drt-dispatch
* transit-capacity-passenger-loading-and-    | * size-battery-electric-bus-fleet-and-
  pass-up-dynamics                           |   chargers
* transit-network-design-and-frequency-      |
  setting                                    |
* urban-freight-delivery-tours-container-    |
  semantics-and-policy-levers                |

[Calibration, estimation, and experimental design]
Knowledge pages (22)                         | Procedural skills (21)
---------------------------------------------+---------------------------------------------
* av-penetration-and-carfollowing-model-     | * build-macroscopic-fundamental-diagram
  mechanism                                  |
* car-following-parameter-calibration-and-   | * build-rolling-horizon-traffic-forecast-
  identifiability                            |   with-state-warm-start
* demand-arrival-process-and-unsignalized-   | * calibrate-car-following-parameters-
  capacity                                   |   against-field-targets
* driver-desired-speed-and-speed-            | * calibrate-demand-with-routesampler
  enforcement-evaluation                     |
* geh-statistic                              | * calibrate-desired-speed-and-evaluate-
                                             |   speed-enforcement
* global-sensitivity-analysis-and-parameter- | * calibrate-flow-with-in-simulation-
  interactions-in-sumo                       |   calibrator
* heavy-vehicle-passenger-car-equivalent-in- | * calibrate-lane-changing-parameters-at-a-
  sumo                                       |   freeway-diverge
* kinematic-wave-theory-validity-across-car- | * calibrate-motorist-yielding-and-select-
  following-models                           |   midblock-crossing-treatment
* lane-change-model-calibration-and-         | * characterize-pedestrian-flow-and-striping-
  identifiability-at-a-diverge               |   model-artifacts
* macroscopic-fundamental-diagram            | * design-count-station-locations-for-od-
                                             |   estimation
* motorist-yielding-calibration-and-         | * emulate-and-evaluate-partial-sensor-
  midblock-crossing-treatment-selection      |   traffic-state-estimation
* od-matrix-estimation-and-                  | * estimate-od-matrix-with-odme
  underdetermination                         |
* pedestrian-flow-theory-and-striping-model- | * estimate-stochastic-freeway-capacity-and-
  artifacts                                  |   breakdown-probability
* routesampler                               | * measure-av-penetration-effect-on-
                                             |   bottleneck-capacity
* sensor-location-design-for-od-estimation   | * measure-heavy-vehicle-passenger-car-
                                             |   equivalent
* simulation-based-optimization-under-noise- | * measure-saturation-flow-and-validate-
  and-seed-overfitting                       |   webster-method
* state-serialization-and-rolling-horizon-   | * model-demand-arrival-process-and-its-
  traffic-forecasting                        |   effect-on-capacity-and-delay
* stochastic-freeway-capacity-and-breakdown- | * optimize-under-simulation-noise-with-a-
  probability                                |   fixed-budget
* sumo-calibrator                            | * quantify-sumo-run-to-run-variability
* sumo-stochastic-variability-and-           | * screen-and-decompose-sumo-parameter-
  replication-design                         |   sensitivity
* traffic-state-estimation-sensor-bias-and-  | * validate-kinematic-wave-theory-across-car-
  sensing-tradeoffs                          |   following-models
* webster-method                             |

[Impact analysis, validation, and visualization]
Knowledge pages (13)                         | Procedural skills (13)
---------------------------------------------+---------------------------------------------
* accessibility-measurement-and-transport-   | * analyze-intersection-air-quality-hot-
  equity                                     |   spots-from-microsimulation
* georeferencing-sumo-output-and-            | * analyze-intersection-safety-with-ssm
  cartographic-fidelity                      |
* harmonoise-traffic-noise-modeling          | * analyze-traffic-noise-with-harmonoise
* hcm-control-delay-vs-sumo-delay-metrics    | * appraise-project-alternatives-with-
                                             |   benefit-cost-analysis
* intersection-air-quality-hot-spot-analysis | * evaluate-multimodal-accessibility-and-
                                             |   equity
* network-safety-screening-and-crash-        | * generate-hcm-los-report-and-validate-
  prediction                                 |   against-microsimulation
* spatial-congestion-heatmap-with-plot-net-  | * measure-travel-time-reliability-with-
  dump                                       |   simulated-days
* sumo-plotting-tools                        | * publish-georeferenced-and-animated-results
* surrogate-safety-measures                  | * screen-network-safety-with-spf-and-
                                             |   empirical-bayes
* teleport-artifacts-and-gridlock-           | * simulate-fleet-emissions
  resolution-validity                        |
* transport-economic-appraisal-from-         | * validate-congested-scenario-results-
  microsimulation                            |   against-teleport-artifacts
* travel-time-reliability-metrics-in-sumo    | * visualize-network-congestion-heatmap
* vehicle-emissions-modeling                 | * visualize-trajectories-and-timeseries

\end{lstlisting}

\section{Supplementary Traces for Case Studies}
\label{app:learning_traces}

The following listings provide condensed, lightly normalized traces for the cases discussed in Section~\ref{sec:case_studies}. They preserve the relevant retrieval, execution, criticism, and memory-update decisions while omitting command-level output, repeated status messages, and long numerical tables.

\subsection{Retrieval and Task Decomposition for Task PP-T4-3-V2}

Listing~\ref{lst:pp_retrieval_trace} shows how memory retrieval supplied a heterogeneous set of components for an infrastructure-planning problem. The decomposition was generated after retrieval; it was not stored as a fixed workflow.

\begin{lstlisting}[caption={Abridged retrieval and decomposition trace for PP-T4-3-V2.},
                   label={lst:pp_retrieval_trace},
                   basicstyle=\ttfamily\scriptsize,
                   breaklines=true,
                   columns=fullflexible]
Retrieved procedural memory (8):
  1. create-grid-network
  2. convert-od-matrix-to-trips
  3. run-simulation
  4. simulate-multimodal-transit
  5. evaluate-multimodal-accessibility-and-equity
  6. appraise-project-alternatives-with-benefit-cost-analysis
  7. solve-budget-constrained-network-design-problem
  8. build-pedestrian-crossings-and-phasing

Retrieved semantic memory (2):
  9. discrete-network-design-and-project-interaction
 10. accessibility-measurement-and-transport-equity

Task-specific decomposition:
  - construct a 5 x 5 grid with a river constraint and five projects;
  - generate and route OD demand;
  - enumerate feasible portfolios and select 14 by D-optimal design;
  - simulate the selected portfolios;
  - fit a surrogate and predict the feasible portfolio space;
  - identify the Pareto frontier and conduct stress tests;
  - evaluate accessibility, equity, and the 5% loss floor;
  - monetize benefits and report distributional outcomes.

No single retrieved item specifies this end-to-end workflow. The action
agent composes it from the retrieved procedures and knowledge pages.
\end{lstlisting}

\subsection{Curriculum Revision After a Failed Task}

Listing~\ref{lst:curriculum_revision_trace} preserves the transition from the original shared-micromobility hypothesis to a stricter curriculum principle and then to a verified trajectory-reconstruction gap.

\begin{lstlisting}[caption={Abridged curriculum-level learning trace.},
                   label={lst:curriculum_revision_trace},
                   basicstyle=\ttfamily\scriptsize,
                   breaklines=true,
                   columns=fullflexible]
Initial gap:
  Memory contains dispatched fleets, scheduled transit, and private
  vehicles in finite parking facilities, but no procedure for modeling
  a fixed fleet of bicycles that wait at stations and are reused by
  different travelers.

Initial premise:
  Assume that SUMO lacks this behavior and construct an alternative.

Outer attempt 1:
  Initial conclusion: bicycle trips create bicycles automatically, so
  SUMO cannot restrict service to a fixed shared fleet.
  Corrected finding: define a fixed set of named bicycles and hold them
  at person-triggered stops. Additional travelers then remain unserved
  while every bicycle is in use.

Outer attempt 2:
  Initial conclusion: each traveler must be assigned a particular
  bicycle before the simulation starts.
  Corrected finding: lines="ANY" lets SUMO assign any eligible bicycle
  when the traveler boards. The earlier test failed because its
  candidate routes were invalid and route errors were suppressed.

Outer attempt 3:
  Initial conclusion: the same bicycle cannot first wait at a station
  and later carry a traveler.
  Corrected finding: appropriate capacity and finite stop-extension
  settings allow one bicycle to alternate between the two roles.

Ingestion decision:
  Create no skill whose justification depends on the rejected premise.
  Create station-based-shared-micromobility-in-sumo as a semantic page.
  Correct knowledge about stopinfo fields and parking occupancy.

Curriculum principle learned:
  "Memory does not cover X" does not imply "SUMO cannot do X."
  Establish absence adversarially before proposing a replacement.
  Characterization, validation, comparison, and integration are also
  legitimate sources of novelty.

Next task:
  Reconstruct SUMO routes and an OD matrix from sparse GPS probes.
  First verify that tracemapper.py and route2OD.py exist; then confirm
  that memory covers only fixed-point count-based reconstruction.
\end{lstlisting}

\subsection{Predictive-Control Skill Acquisition}

Listing~\ref{lst:predictive_learning_trace} summarizes the complete curriculum--execution--criticism--ingestion--maintenance sequence for the predictive-control learning episode.

\begin{lstlisting}[caption={Abridged end-to-end trace for predictive-control learning.},
                   label={lst:predictive_learning_trace},
                   basicstyle=\ttfamily\scriptsize,
                   breaklines=true,
                   columns=fullflexible]
CURRICULUM
Gap: existing signal controllers are offline-static or reactive;
none optimizes a sequence of future decisions using predicted arrivals.
Task: compare two predictive signal controllers: an analytical
dynamic-programming controller and an MPC controller that evaluates
candidate decisions through short SUMO rollouts. Validate the arrival
forecasts separately; compare detector-based prediction with
non-informative and perfect-information controls; repeat the experiment
across demand levels and arrival patterns using multiple random seeds;
and test whether predictive control coordinates two successive signals.

RETRIEVAL
Skills:
  implement-maxpressure-traci-controller
  build-rolling-horizon-traffic-forecast-with-state-warm-start
  design-actuated-signal-detector-placement-and-fault-tolerance
  model-demand-arrival-process-and-its-effect-on-capacity-and-delay
  optimize-signals-by-tlscycleadaptation
  quantify-sumo-run-to-run-variability
Pages:
  state-serialization-and-rolling-horizon-traffic-forecasting
  actuated-signal-detector-design-and-fault-tolerance
  demand-arrival-process-and-unsignalized-capacity
  coordinated-adaptive-signal-control-detector-bias-and-transition-cost

OUTER ATTEMPT 1
The action agent produces executable controllers and simulation results.
Critic verdict: success=false.
Successful execution alone does not validate the experimental design.
The critic requests independent checks of state restoration, sequence
optimization, prediction-null equivalence, controller settings, and
the scope of the reported conclusions.

OUTER ATTEMPT 2
A fresh action-agent invocation receives the accumulated critic
feedback, but not the command-level debugging history from the first
invocation. It tests each controller across plausible settings, checks
the dynamic-programming solver against exhaustive enumeration,
separates predictor and optimizer effects, and repeats critical cases.
Critic verdict: success=true after nine independent checks.

INGESTION
New skill:
  implement-predictive-rolling-horizon-signal-control
  (+3 bundled scripts)
Updated skills:
  build-rolling-horizon-traffic-forecast-with-state-warm-start
  optimize-under-simulation-noise-with-a-fixed-budget
New page:
  value-of-anticipation-in-predictive-signal-control
Updated page:
  state-serialization-and-rolling-horizon-traffic-forecasting

MEMORY LINT
Check ten changed artifacts against all existing skills and pages;
parse 3/3 new scripts; resolve all links; synchronize page summaries;
confirm that predictive and max-pressure control remain separate
because one is anticipatory and the other reactive; confirm that
count-station design and OD estimation remain separate because one
selects sensors and the other estimates demand. No repair is required.
\end{lstlisting}

\subsection{Compositional Map-Matching Skill Acquisition}

Listing~\ref{lst:map_matching_learning_trace} shows how the action agent composed existing procedures into a new skill and, during reuse, discovered and repaired an error in one of those procedures.

\begin{lstlisting}[caption={Abridged composition and ingestion trace for GPS map matching.},
                   label={lst:map_matching_learning_trace},
                   basicstyle=\ttfamily\scriptsize,
                   breaklines=true,
                   columns=fullflexible]
TASK
Run a SUMO scenario and retain each vehicle's exact route as ground
truth. Sample the simulated vehicle positions to create synthetic GPS
records, then vary the time between records, positional error, the
fraction of vehicles observed, and the proportion of missing records.
Use tracemapper.py to infer a route for each observed vehicle and
route2OD.py to aggregate the inferred routes into an OD matrix. Compare
the inferred edges, complete routes, and OD flows with the ground truth.

COMPOSITION
Reuse network-import, demand-reconstruction, replication, and output-
analysis skills. Bundle new scripts for probe generation, matching,
scoring, and OD aggregation.

DEPENDENCY TEST AND REPAIR
The reused load-osm-network wrapper passes --bbox and its coordinates
as separate arguments. For western-hemisphere bounding boxes, the
negative longitude is parsed as another command-line option and the
download fails. The action agent changes the call to the single-token
form --bbox=..., then verifies the repaired wrapper.

CONSISTENT INGESTION
New skill:
  map-match-gps-traces-to-reconstruct-demand (+4 bundled scripts)
Repaired skill and script:
  load-osm-network (use the single-token --bbox=... form)
New page:
  gps-map-matching-and-probe-demand-reconstruction
Updated page:
  geh-statistic (independent evidence on fit versus OD recovery)
\end{lstlisting}

\section{Memory-Guided Congestion--Safety Cross-Check (OA-T3)}
\label{app:oa_t3}

This and the following two appendices trace memory-guided inference under the complete SimSkill condition. Each case follows the same sequence: retrieval of procedural skills and semantic knowledge, action-agent execution and critic verification, and production of a final result with reproducible artifacts. All three tasks were executed with DeepSeek-V4-Pro in test mode. Memory was available for retrieval, but ingestion was disabled, so the evaluated memory snapshot remained fixed across tasks. Each episode was accepted by the in-loop critic and subsequently passed the independent artifact-based verification.

\subsection{Task and Memory Use}

OA-T3 asks the agent to build a spatial congestion heat map for a $4\times4$-block signalized grid with 200~m spacing and heavy demand of 1,000~veh/h at each inbound fringe edge, and then determine whether the most congested edges coincide with the locations reporting the most surrogate-safety-measure (SSM) conflicts. The task is compositional: network and demand generation, simulation, output visualization, safety-device configuration, spatial matching, and statistical comparison must operate on one consistent scenario.

Retrieval returned six procedural skills. \texttt{create-grid-network}, \texttt{generate-random-trips}, and \texttt{run-simulation} supplied network construction, demand generation, and reproducible execution. \\ 
\texttt{visualize-network-congestion-heatmap} supplied interval-based heat-map generation; \\
\texttt{analyze-intersection-safety-with-ssm} supplied SSM configuration, conflict parsing, and spatial assignment; and \texttt{analyze-simulation-outputs} supplied output-integrity checks. The two semantic pages shaped analytical decisions rather than execution. \\
\texttt{spatial-congestion-heatmap-with-plot-net-dump} guided the choice of lane occupancy as the congestion metric, while \texttt{surrogate-safety-measures} provided the meaning of the safety indicators, thresholds, and conflict types.

\subsection{Memory-Guided Inference and Critic Revision}

The action agent composed the retrieved artifacts into a workflow that generated one common simulation, produced twelve interval-level congestion maps, assigned SSM conflicts to directed edges, and compared the resulting congestion and safety rankings. This workflow was not copied from a single skill: the LLM connected network, simulation, visualization, safety-analysis, and validation procedures to satisfy the cross-domain task.

The episode used all three outer action--critic attempts. The critic rejected the first attempt after detecting an incorrect demand calculation and a silent conflict-parser failure. The second action-agent attempt repaired both defects, but the critic found that the written interpretation still disagreed with the output files and omitted important limitations. The third attempt reconciled the report with the artifacts, corrected the hotspot description, and stated the insertion and ranking limitations explicitly. The trace therefore shows that SimSkill inference continues beyond successful program execution until the critic accepts both the artifacts and the claims derived from them.

\subsection{Verified Result and Output}

Table~\ref{tab:oa_t3_results} summarizes the accepted evidence. No edge appeared in both top-ten lists, although the full 120-edge rankings had a Spearman correlation of 0.748. The result therefore distinguishes broad rank association from agreement among the most critical locations.

\begin{table*}[t]
    \centering
    \small
    \setlength{\tabcolsep}{5pt}
    \caption{Accepted OA-T3 congestion--safety cross-check. SSM counts are raw ego--foe records.}
    \label{tab:oa_t3_results}
    \begin{tabular}{p{0.22\textwidth}p{0.24\textwidth}p{0.44\textwidth}}
        \hline
        Quantity & Value & Interpretation \\
        \hline
        Demand realization & 20,000 scheduled; 3,619 inserted; 2,304 completed & The intended demand severely oversaturated the network; the realized pattern represents only the vehicles that entered. \\
        SSM records & 37,721 raw; approximately 18,860 distinct events & Each physical interaction was recorded twice with ego and foe exchanged; rankings are unaffected. \\
        Hotspot overlap & Top five: 0/5; top ten: 0/10 & The highest-congestion and highest-conflict locations were disjoint. \\
        Full-ranking association & $\rho=0.748$ overall & Bulk agreement did not imply agreement at the top of the rankings. \\
        Type-specific association & Rear-end: 0.652; crossing: 0.316; merging: $-0.024$ & The association was driven primarily by rear-end interactions. \\
        \hline
    \end{tabular}
\end{table*}

\begin{figure}
    \centering
    \includegraphics[width=0.94\textwidth]{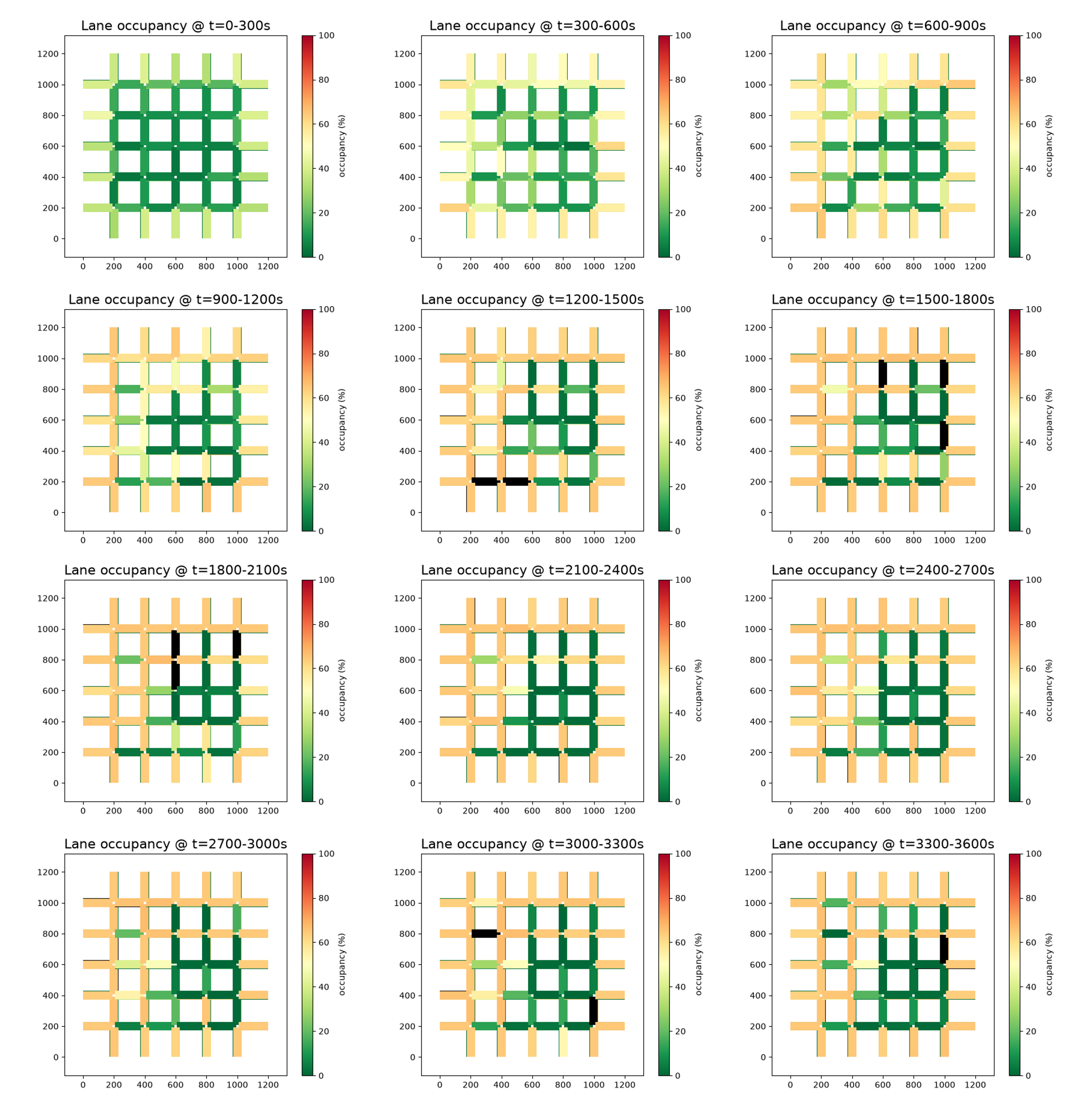}
    \caption{OA-T3 lane-occupancy heat maps for twelve consecutive 300~s intervals. The maps were generated from edge-data output using the retrieved congestion-visualization procedure and semantic guidance on metric selection. Their interval maxima produced the congestion ranking subsequently compared with spatially assigned SSM conflicts.}
    \label{fig:oa_t3_result}
\end{figure}

The final output included reproducible build-and-analysis scripts, edge-data and SSM files from the common scenario, a combined edge-ranking table, the corrected report, and the twelve heat maps shown in Figure~\ref{fig:oa_t3_result}. The maps instantiate the retrieved visualization procedure and the congestion metric selected from semantic memory; their rankings were then compared with conflicts parsed using the retrieved safety procedure. Within the stated limitations, the verified result is that a congestion heat map cannot substitute for direct safety-hotspot analysis in this scenario.

\section{Memory-Guided Time-Dependent OD Inference (DG-T4-3-V2)}
\label{app:dg_t4_3_v2}

\subsection{Task and Memory Use}

DG-T4-3-V2 is a demand-inference task on a $3\times3$ grid with six boundary traffic analysis zones. It asks for posterior distributions over all 30 directed origin--destination (OD) pairs in eight 15-minute intervals, yielding 240 nonnegative flows, from only 24 screenline counts and sparse Bluetooth travel times. Holdout rows must be excluded from fitting and model selection, the result must quantify non-identifiability, and the simulation budget is fixed at 15 designated SUMO executions. This is a useful test of compositional inference because the stored library contained relevant foundations but no end-to-end procedure for likelihood-free, time-dependent OD estimation.

Retrieval returned five procedural skills. \texttt{create-grid-network} and \texttt{run-simulation} supplied the network and reproducible execution foundation. \texttt{estimate-od-matrix-with-odme} supplied the assignment-matrix logic, a deterministic comparator, and diagnostics separating count fit from OD recovery. \texttt{optimize-under-simulation-noise-with-a-fixed-budget} contributed the general strategy of replacing repeated SUMO calls with a task-specific surrogate and maintaining an auditable record of all executions permitted by the 15-run simulation budget. \texttt{quantify-sumo-run-to-run-variability} guided the separation of training and validation seeds.

The two retrieved knowledge pages shaped interpretation and validation. \\
\texttt{od-matrix-estimation-and-underdetermination} led the agent to report uncertainty and non-identifiable contrasts instead of presenting one fitted OD matrix as ground truth. \texttt{geh-statistic} supplied the count goodness-of-fit measure and its limitations. Memory provided relevant OD-estimation principles and validation criteria, but none of the retrieved skills implemented simulation-based Bayesian inference for time-varying OD demand. The action agent therefore constructed a task-specific Approximate Bayesian Computation with Sequential Monte Carlo (ABC--SMC) workflow.

\subsection{Memory-Guided Inference and New Construction}

The 240 time-dependent OD flows could not be estimated independently from the sparse observations. The action agent therefore represented them using a smaller set of parameters describing the spatial demand pattern and its variation over time. The retrieved OD Matrix Estimation (ODME) skill supplied the assignment logic and a deterministic least-squares baseline, while the semantic memory motivated explicit uncertainty and identifiability analysis. The Bayesian method and the ODME baseline used the same routes for a fair comparison, and holdout observations were reserved exclusively for final validation. Because memory contained no procedure for likelihood-free, time-dependent OD inference, the agent implemented an ABC--SMC workflow specifically for this task. The fixed-budget skill directly shaped execution. Twelve SUMO runs trained a physics-informed surrogate; three independent posterior populations then operated entirely on the cached surrogate without calling SUMO; and two posterior representatives plus the ODME comparator consumed the final three runs. The resulting ledger contained 15 designated simulations.

The episode was accepted in its first outer action--critic attempt. The critic independently verified the holdout masks, the 15-run ledger, the absence of SUMO calls during posterior inference, the comparator, and the required convergence and predictive diagnostics. It also found five minor discrepancies between reported numbers and stored artifacts; these were corrected in the report without changing the code or rerunning the experiment. This case shows how retrieved memory can structure a novel solution while the critic enforces the task's evidential requirements.

\subsection{Verified Result and Output}

The accepted output satisfied the simulation budget, convergence checks, and holdout criterion, as summarized in Table~\ref{tab:dg_t4_3_v2_results}. The results also reflected the limitations highlighted by semantic memory: aggregate demand was better identified than individual OD flows, because multiple OD matrices could reproduce similar screenline counts. Moreover, predictions from the surrogate did not always agree with the final SUMO simulations.

\begin{table*}[t]
    \centering
    \small
    \caption{Accepted DG-T4-3-V2 inference and verification results.}
    \label{tab:dg_t4_3_v2_results}
    \begin{tabular}{p{0.20\textwidth}p{0.26\textwidth}p{0.45\textwidth}}
        \hline
        Verification target & Result & Interpretation \\
        \hline
        Simulation budget & 12 training runs and 3 final runs & 15 SUMO executions; posterior inference itself used no additional SUMO calls. \\
        Population agreement & ESS 7,628--7,647 of 8,000; maximum $\widehat{R}=1.007$ & The three independent ABC--SMC populations produced closely aligned posteriors. \\
        Holdout validation &
        All 6 reserved observations fell within their 90\% predictive intervals &
        The resulting 100\% coverage exceeded the required 80\%; the reserved observations were used only for final evaluation. \\
        Calibration-count fit &
        Mean GEH of 5.73; 48\% of fitted counts below 5 &
        Agreement was weaker during peak periods because the simulated boundary connectors could not reproduce the highest observed flows. \\
        Uncertainty in OD estimates &
        90\% credible intervals for all 240 flows, with additional tests of indistinguishable OD combinations &
        The observations constrained aggregate demand more strongly than individual OD flows; the results were therefore reported as distributions rather than as a uniquely determined OD matrix. \\
        \hline
    \end{tabular}
\end{table*}

\begin{figure}
    \centering
    \includegraphics[width=0.98\textwidth]{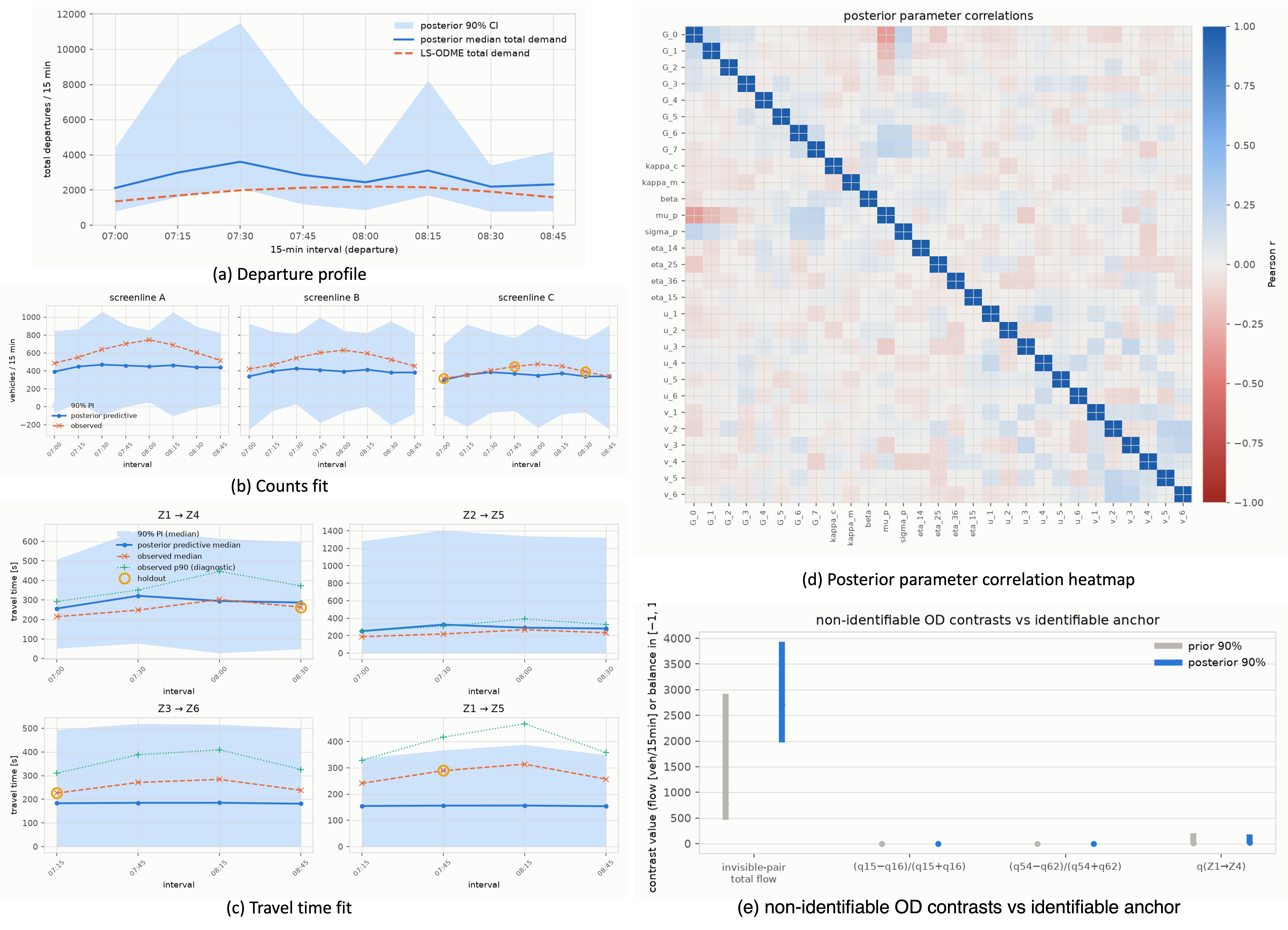}
    \caption{DG-T4-3-V2 posterior diagnostics and results. The left column shows total interval demand, posterior-predictive screenline counts, and travel times. The right column shows posterior parameter correlations and prior-to-posterior changes in selected non-identifiable OD contrasts. Together, the panels distinguish fit to observed aggregates from uncertainty in the underlying OD cells.}
    \label{fig:dg_t4_3_v2_result}
\end{figure}

The final artifacts included the network and signal files, ten task scripts, the 15-run and discarded-run ledgers, the fitted surrogate, posterior samples and metadata, a CSV containing credible intervals for all 240 flows, the deterministic ODME comparator, summary JSON files, and five diagnostic figures. Figure~\ref{fig:dg_t4_3_v2_result} combines the principal outputs: demand and posterior-predictive checks, parameter correlations, and non-identifiable contrasts. It therefore makes visible both the inferred result and the caution supplied by semantic memory about underdetermination.

\section{Procedural Transfer to Shared E-Scooter Operations (MT-T4-4-V2)}
\label{app:mt_t4_4_v2}

\subsection{Task and Selective Memory Use}

MT-T4-4-V2 compares no rebalancing, scheduled truck rebalancing, and incentive-based user rebalancing for a docked shared e-scooter system under operational, environmental, and equity constraints. No semantic page about shared scooters or micromobility operations was present in the memory. The case therefore shows how SimSkill can transfer procedural memory while constructing missing domain-specific logic.

Retrieval returned eight procedural skills, six of which shaped the solution. \texttt{create-grid-network} and \texttt{run-simulation} supplied network construction and SUMO validation; \\ 
\texttt{model-dedicated-bicycle-lane-infrastructure} informed the protected-lane representation; \\
\texttt{model-urban-freight-delivery-tours} supplied a multi-stop routing method for the rebalancing truck and a procedure for calculating the distance it traveled; \texttt{simulate-fleet-emissions} supplied the emissions-reporting structure; and \texttt{evaluate-multimodal-accessibility-and-equity} guided reporting by origin-area income label. The agent also inspected \texttt{get-vehicles-state} and \\
\texttt{set-vehicle-state}, but did not use them because the selected event-driven formulation required no TraCI state-transfer loop. Retrieval thus provided candidate capabilities rather than commands that had to be applied mechanically.

\subsection{Memory-Guided Inference and Task-Specific Extension}

The action agent used the retrieved procedures for network construction and route-time validation, while implementing scooter inventory, battery state, walking, docking, truck movement, and incentive decisions in a task-owned discrete-event engine. This division illustrates adaptive skill use: stored procedures supplied reliable components, whereas the agent constructed the domain-specific mechanism absent from memory.

The agent then executed paired policy comparisons under four base perturbations and two rain perturbations, producing 18 policy--scenario--seed runs. Within the action agent's inner loop, it corrected four local defects involving metric calculation, incentive logic, report generation, and route validation. The critic subsequently reran the complete workflow, obtained byte-identical outputs, checked every explicit task requirement, and accepted the first outer attempt.

\subsection{Verified Result and Output}

Table~\ref{tab:mt_t4_4_v2_results} reports the base-scenario means. Scheduled truck rebalancing raised served share from 58.8\% without rebalancing to 95.8\%; incentives raised it to 75.1\%. The largest distributional change occurred at low-income-origin stations, where served share rose from 40.9\% to 95.7\% under the truck policy. The truck added 10.7~km of combustion-vehicle travel and 2.13~kg of CO$_2$, while the incentive policy incurred no truck travel but paid \$94.0 per simulated day on average. Neither full-dock failures nor sidewalk conflicts occurred in the abstract network.

\begin{table*}[t]
    \centering
    \small
    \setlength{\tabcolsep}{8pt}
    \caption{MT-T4-4-V2 base-scenario results, averaged over four paired perturbation seeds.}
    \label{tab:mt_t4_4_v2_results}
    \begin{tabular}{p{0.41\textwidth}rrr}
        \hline
        Metric & None & Truck & Incentive \\
        \hline
        Served share (\%) & 58.8 & 95.8 & 75.1 \\
        Served share, low-income origins (\%) & 40.9 & 95.7 & 64.7 \\
        Served share, high-income origins (\%) & 94.1 & 95.8 & 95.8 \\
        Total motorized VKT (km) & 43.8 & 82.5 & 75.2 \\
        Truck VKT (km) & 0.0 & 10.7 & 0.0 \\
        Truck CO$_2$ (kg) & 0.00 & 2.13 & 0.00 \\
        Incentive cost (USD) & 0.0 & 0.0 & 94.0 \\
        \hline
    \end{tabular}
\end{table*}

The policy ordering remained stable in the rain sensitivity case, with served shares of 87.6\%, 98.9\%, and 97.8\% for no rebalancing, truck rebalancing, and incentives, respectively. Considering availability, equity, VKT, emissions, and cost together, the action agent recommended the truck policy as the primary strategy and incentives as a rebalancing complement without truck travel.

\begin{figure}
    \centering
    \includegraphics[width=0.98\textwidth]{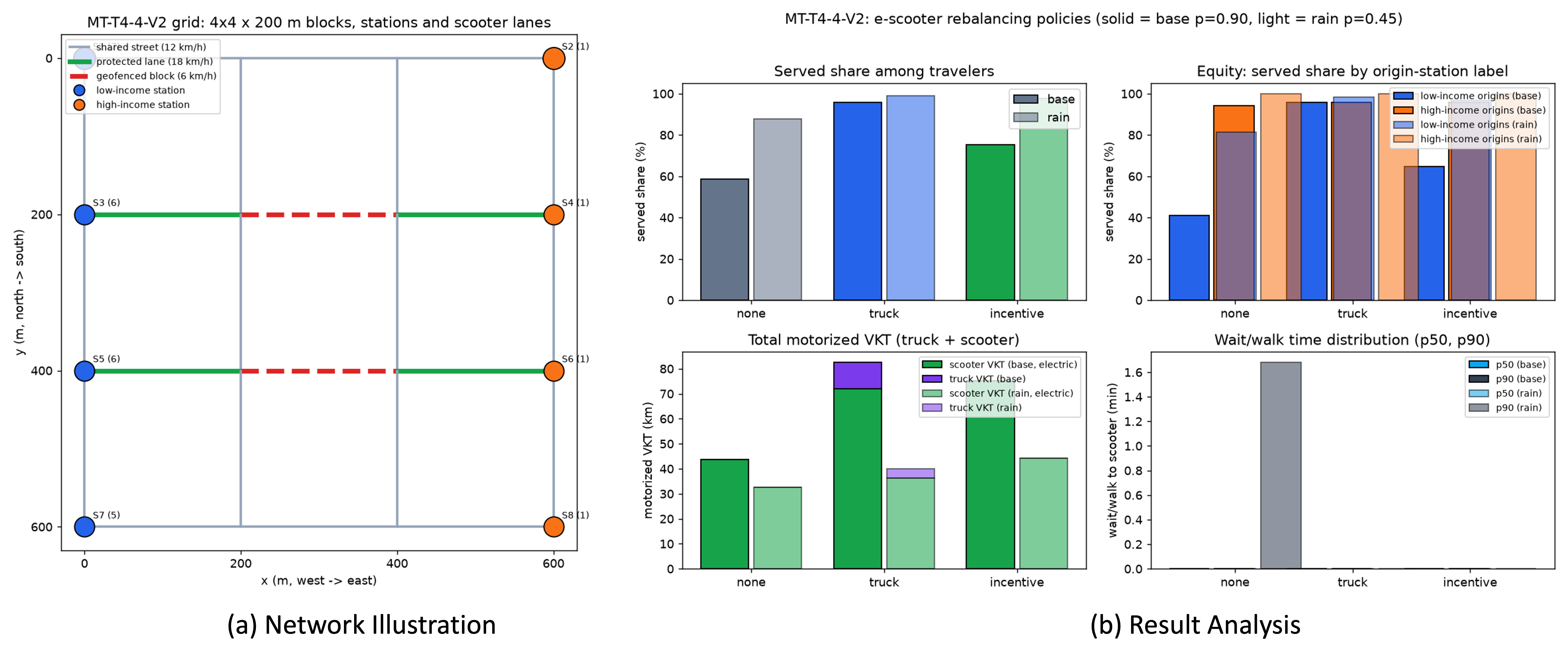}
    \caption{MT-T4-4-V2 network and policy outcomes. The left panel shows station locations, income-area labels, protected scooter lanes, and geofenced low-speed blocks. The right panels compare served share, served share by origin-station label, total motorized VKT, and wait/walk time for no rebalancing, truck rebalancing, and incentive rebalancing under base and rain demand.}
    \label{fig:mt_t4_4_v2_result}
\end{figure}

The final output included five task scripts, the SUMO network and specification, route-time validation, machine-readable aggregate and per-seed results, request-level outcome files, a written report, and two figures. Figure~\ref{fig:mt_t4_4_v2_result} combines the network representation derived from the retrieved infrastructure procedures with the policy, equity, and VKT outcomes produced by the task-owned engine.

\subsection{Cross-Case Interpretation}

The three cases illustrate complementary forms of memory-guided inference. OA-T3 composed mature procedural and semantic artifacts and required repeated critic-guided correction. DG-T4-3-V2 used memory as methodological scaffolding for a likelihood-free estimator that had to be constructed in task-owned code. MT-T4-4-V2 selectively transferred procedural skills despite having no matching semantic page and set aside retrieved skills that were unnecessary. Across all three cases, the action agent selected, adapted, and composed memory into a task-specific workflow; the critic verified the artifacts and claims; and the final output preserved scripts, data, results, and figures for reproduction. Because all tasks ran in test mode, none could alter the memory available to subsequent benchmark tasks.

\end{document}